\documentclass{article}

\usepackage[T1]{fontenc}

\usepackage{tgtermes}
\usepackage{amsmath}
\usepackage{amssymb}
\usepackage{scalefnt,letltxmacro}
\LetLtxMacro{\oldtextsc}{\textsc}
\renewcommand{\textsc}[1]{\oldtextsc{\scalefont{1.10}#1}}
\usepackage[scaled=0.92]{PTSans}
\usepackage{inconsolata}
\usepackage{mathbbol}

\usepackage[usenames,dvipsnames]{xcolor}
\definecolor{shadecolor}{gray}{0.9}

\usepackage[final, expansion=alltext]{microtype}
\usepackage[english]{babel}
\usepackage{afterpage}
\usepackage{framed}
\usepackage{nicefrac}

{\endMakeFramed}

\DeclareRobustCommand{\parhead}[1]{\textbf{#1}~}

\usepackage{graphicx}
\usepackage[labelfont=bf]{caption}
\usepackage[format=hang]{subcaption}

\usepackage{booktabs}
\usepackage{multirow}
\usepackage{longtable}
\usepackage{etoolbox,siunitx}
\robustify\bfseries
\usepackage{natbib}
\usepackage{bibunits}

\usepackage{listings}
\usepackage{fancyvrb}
\fvset{fontsize=\normalsize}
\usepackage{algorithm}
\usepackage{algorithmic}

\usepackage[colorlinks, linktoc=all]{hyperref}
\usepackage[all]{hypcap}
\hypersetup{citecolor=Violet}
\hypersetup{linkcolor=black}
\hypersetup{urlcolor=MidnightBlue}

\usepackage[nameinlink]{cleveref}

\usepackage[acronym,smallcaps,nowarn]{glossaries}
\glsdisablehyper{}

\usepackage{listings}
\lstdefinestyle{mystyle}{
    commentstyle=\color{OliveGreen},
    numberstyle=\tiny\color{black!60},
    stringstyle=\color{BrickRed},
    basicstyle=\ttfamily\scriptsize,
    breakatwhitespace=false,
    breaklines=true,
    captionpos=b,
    keepspaces=true,
    numbers=none,
    numbersep=5pt,
    showspaces=false,
    showstringspaces=false,
    showtabs=false,
    tabsize=2
}
\DeclareRobustCommand{\mb}[1]{\ensuremath{\mathbf{\boldsymbol{#1}}}}

\newtheorem{theorem}{Theorem}

\newtheorem{assumption}{Assumption}
\newcommand\independent{\protect\mathpalette{\protect\independenT}{\perp}}
\def\independenT#1#2{\mathrel{\rlap{$#1#2$}\mkern2mu{#1#2}}}

\renewcommand{\mid}{~\vert~}
\newcommand{\prm}{\:;\:}

\newcommand{\mbone}{\mb{1}}

\newcommand{\mbtheta}{\mb{\theta}}

\newcommand{\mbbeta}{\mb{\beta}}

\newcommand\dif{\mathop{}\!\mathrm{d}}

\newcommand{\E}{\mathbb{E}}

\newcommand{\bbR}{\mathbb{R}}

\newcommand{\cN}{\mathcal{N}}

\newcommand{\Gam}{\textrm{Gam}}

\newacronym{ELBO}{elbo}{evidence lower bound}
\newacronym{GMM}{gmm}{Gaussian mixture model}
\newacronym{KL}{kl}{Kullback-Leibler}
\newacronym{LDA}{lda}{latent Dirichlet allocation}
\newacronym{SVI}{svi}{stochastic variational inference}

\newacronym{MLE}{mle}{maximum likelihood estimate}
\newacronym{MAP}{map}{maximum-a-posterior}
\newacronym{MCMC}{mcmc}{Markov chain Monte Carlo}

\newacronym{LBFGS}{l-bfgs}{limited-memory Broyden-Fletcher-Goldfarb-Shanno}
\newacronym{ADVI}{advi}{automatic differentiation variational inference}
\newacronym{NUTS}{nuts}{No-U-Turn sampler}

\newacronym{GLM}{glm}{generalized linear model}

\newacronym{IF}{if}{influence function}

\newacronym{PF}{pf}{Poisson factorization}

\newacronym[\glsshortpluralkey={rpm}]{RPM}{rpm}{reweighted probabilistic model}

\newacronym{DPMM}{dpmm}{Dirichlet process mixture model}

\usepackage{tikz}
\usetikzlibrary{bayesnet}
\usepackage{pgfplots}
\pgfplotsset{compat=newest}
\pgfplotsset{plot coordinates/math parser=false}
\usepgfplotslibrary{statistics}

\pgfdeclarelayer{edgelayer}
\pgfdeclarelayer{nodelayer}
\pgfsetlayers{edgelayer,nodelayer,main}

\definecolor{hexcolor0xbfbfbf}{rgb}{0.749,0.749,0.749}

\tikzset{>=latex}
\tikzstyle{none}   = [inner sep=0pt]
\tikzstyle{line}   = [ thick, -, shorten <=1pt, shorten >=1pt ]
\tikzstyle{arrow}  = [ thick,  ->, shorten <=1pt, shorten >=1pt ]
\tikzstyle{ardash} = [ thick dotted, ->, shorten <=1pt, shorten >=1pt ]

\tikzstyle{empty}=[circle,opacity=0.0,text opacity=1.0,minimum width=4pt,minimum height=4pt]
\tikzstyle{box}=[rectangle,fill=White,draw=Black]
\tikzstyle{filled}=[circle,fill=hexcolor0xbfbfbf,draw=Black]
\tikzstyle{hollow}=[circle,fill=White,draw=Black]
\tikzstyle{param}=[rectangle,fill=Black,draw=Black,inner sep=0pt,minimum width=4pt,minimum height=4pt]
\tikzstyle{paramhollow}=[rectangle,fill=White,draw=Black,inner sep=0pt,minimum
width=4pt,minimum height=4pt]

\usepackage{times}

\usepackage[accepted]{icml2017}
\icmltitlerunning{Robust Probabilistic Modeling with Bayesian Data Reweighting}

\begin{document}

\twocolumn[
\icmltitle{Robust Probabilistic Modeling with Bayesian Data Reweighting}

\begin{icmlauthorlist}
\icmlauthor{Yixin Wang}{columbia}
\icmlauthor{Alp Kucukelbir}{columbia}
\icmlauthor{David M.~Blei}{columbia}
\end{icmlauthorlist}

\icmlaffiliation{columbia}{Columbia University, New York City, USA}
\icmlcorrespondingauthor{Yixin Wang}{yixin.wang@columbia.edu}

\icmlkeywords{}
\vskip 0.3in]

\printAffiliationsAndNotice{}

\begin{abstract}
Probabilistic models analyze data by relying on a set of assumptions.
Data that exhibit deviations from these assumptions can undermine
inference and prediction quality. Robust models offer protection
against mismatch between a model's assumptions and reality. We
propose a way to systematically detect and mitigate mismatch of a
large class of probabilistic models. The idea is to raise the
likelihood of each observation to a weight and then to infer both the
latent variables and the weights from data. Inferring the weights
allows a model to identify observations that match its assumptions
and down-weight others. This enables robust inference and improves
predictive accuracy. We study four different forms of mismatch with
reality, ranging from missing latent groups to structure
misspecification. A Poisson factorization analysis of the Movielens
1M dataset shows the benefits of this approach in a practical
scenario.
\end{abstract}

\begin{bibunit}[apa]

\section{Introduction}
\label{sec:introduction}

Probabilistic modeling is a powerful approach to discovering hidden
patterns in data. We begin by expressing assumptions about the class
of patterns we expect to discover; this is how we design a
probability model. We follow by inferring the posterior of the model;
this is how we discover the specific patterns manifest in an observed
data set. Advances in automated inference
\citep{hoffman2014nuts,mansinghka2014venture,kucukelbir2016automatic}
enable easy development of new models for machine learning and
artificial intelligence \citep{ghahramani2015probabilistic}.

In this paper, we present a recipe to robustify probabilistic models.
What do we mean by ``robustify''? Departure from a model's
assumptions can undermine its inference and prediction performance.
This can arise due to corrupted observations, or in general,
measurements that do not belong to the process we are modeling.
Robust models should perform well in spite of such mismatch with
reality.

Consider a movie recommendation system. We gather data of people
watching movies via the account they use to log in. Imagine a
situation where a few observations are corrupted For example, a child
logs in to her account and regularly watches popular animated films.
One day, her parents use the same account to watch a horror movie.
Recommendation models, like \Gls{PF}, struggle with this kind of
corrupted data (see \Cref{sec:movielens}): it begins to recommend
horror movies.

What can be done to detect and mitigate this effect? One strategy is
to design new models that are less sensitive to corrupted data, such
as by replacing a Gaussian likelihood with a heavier-tailed $t$
distribution \citep{huber2011robust,insua2012robust}. Most
probabilistic models we use have more sophisticated structures; these
template solutions for specific distributions are not readily
applicable. Other classical robust techniques act mostly on distances
between observations \citep{huber1973robust}; these approaches
struggle with high-dimensional data. How can we still make use of our
favorite probabilistic models while making them less sensitive to the
messy nature of reality?

\textbf{Main idea.} We propose \glspl{RPM}. The idea is simple.
First, posit a probabilistic model. Then adjust the contribution of
each observation by raising each likelihood term to its own (latent)
weight. Finally, infer these weights along with the latent variables
of the original probability model. The posterior of this adjusted
model identifies observations that match its assumptions; it
downweights observations that disagree with its assumptions.

\begin{figure}[htb]
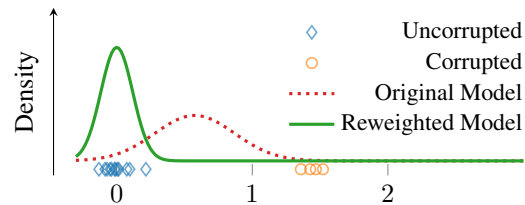

\centering
   
   \include{img/intuition_plot}
   \vspace*{-18pt}
   \caption{Fitting a unimodal distribution to a dataset with corrupted
   measurements. The \gls{RPM} downweights the corrupted observations.}
   \label{fig:intuition_plot}
\end{figure}

\Cref{fig:intuition_plot} depicts this tradeoff. The dataset includes
corrupted measurements that undermine the original model; Bayesian
data reweighting automatically trades off the low likelihood of the
corrupted data near 1.5 to focus on the uncorrupted data near zero.
The \gls{RPM} (green curve) detects this mismatch and mitigates its
effect compared to the poor fit of the original model (red curve).

Formally, consider a dataset of $N$ independent observations $y =
(y_1,\ldots,y_N)$. The likelihood factorizes as a product
$\prod_{n=1}^N
\ell(y_n\mid\beta)$, where $\beta$ is a set of latent variables.
Posit a prior distribution $p_\beta(\beta)$.

Bayesian data reweighting follows three steps:
\begin{enumerate}
  \item Define a probabilistic model $p_\beta(\beta) \prod_{n=1}^N
  \ell(y_n\mid\beta)$.
  \item Raise each likelihood to a positive latent weight $w_n$. Then
  choose a prior on the weights $p_w(w)$, where $w =
  (w_1,\ldots,w_N)$. This gives a reweighted probabilistic model
  (\textsc{rpm})
  \begin{align*}
  p(y,\beta,w)
  &=
  \frac{1}{Z}
  p_\beta(\beta)
  p_w(w)
  \prod_{n=1}^N \ell(y_n\mid\beta)^{w_n},
\end{align*}
where $Z$ is the normalizing factor.

  \item Infer the posterior of both the latent variables $\beta$ and
  the weights $w$, $p(\beta,w\mid y)$.
\end{enumerate}
The latent weights $w$ allow an \gls{RPM} to automatically explore
which observations match its assumptions and which do not. Writing
out the logarithm of the \gls{RPM} gives some intuition; it is equal
(up to an additive constant) to
\begin{align}
  \log p_\beta (\beta) + \log p_w(w) + \sum_n w_n \log \ell(y_n\mid\beta).
  \label{eq:logRPM}
\end{align}
Posterior inference, loosely speaking, seeks to maximize the above
with respect to $\beta$ and $w$. The prior on the weights $p_w(w)$
plays a critical role: it trades off extremely low likelihood terms,
caused by corrupted measurements, while encouraging the weights to be
close to one. We study three options for this prior in
\Cref{sec:weighted}.

How does Bayesian data reweighting induce robustness? First, consider
how the weights $w$ affect \Cref{eq:logRPM}. The logarithm of our
priors are dominated by the $\log w_n$ term: this is
the price of moving $w_n$ from one towards zero. By shrinking $w_n$,
we gain an increase in $w_n \log \ell(y_n\mid\beta)$ while paying a
price in $a\log w_n$. The gain outweighs the price we pay if $\log
\ell(y_n\mid\beta)$ is very negative. Our priors are set to prefer
$w_n$ to stay close to one; an \gls{RPM} only shrinks $w_n$ for very
unlikely (e.g., corrupted) measurements.

Now consider how the latent variables $\beta$ affect
\Cref{eq:logRPM}. As the weights of unlikely measurements shrink, the
likelihood term can afford to assign low mass to those corrupted
measurements and focus on the rest of the dataset. Jointly, the
weights and latent variables work together to automatically identify
unlikely measurements and focus on observations that match the
original model's assumptions.

\Cref{sec:weighted} presents these intuitions in full detail, along
with theoretical corroboration. In \Cref{sec:empirical}, we study
four models under various forms of mismatch with reality, including
missing modeling assumptions, misspecified nonlinearities, and skewed
data. \gls{RPM}s provide better parameter inference and improved
predictive accuracy across these models. \Cref{sec:movielens}
presents a recommendation system example, where we improve on
predictive performance and identify atypical film enthusiasts in the
Movielens 1M dataset.

\textbf{Related work.} Jerzy Neyman elegantly motivates the main idea
behind robust probabilistic modeling, a field that has attracted much
research attention in the past century.

\begin{quotation}
Every attempt to use mathematics to study some real phenomena must
begin with building a mathematical model of these phenomena. Of
necessity, the model simplifies matters to a greater or lesser extent
and a number of details are ignored. [...] The solution of the mathematical problem may be correct and yet it
may be in violent conflict with realities simply because the original
assumptions of the mathematical model diverge essentially from the
conditions of the practical problem considered.
\citep[p.22]{neyman1949problem}.
\end{quotation}

Our work draws on three themes around robust modeling.

The first is a body of work on robust statistics and machine learning
\citep{provost2001robust,song2002robust,NIPS2012_4558,NIPS2014_5428,NIPS2014_5515,NIPS2015_5745}.
These developments focus on
making specific models more robust to imprecise measurements.

One strategy is popular: localization. To localize a probabilistic
model, allow each likelihood to depend on its own ``copy'' of the
latent variable $\beta_n$. This transforms the model into
\begin{align}
  p(y,\beta,\alpha)
  &=
  p_\alpha(\alpha)
  \prod_{n=1}^N \ell(y_n\mid \beta_n)
  p_{\beta}(\beta_n \mid \alpha),
\label{eq:localization}
\end{align}
where a top-level latent variable $\alpha$ ties together all the
$\beta_n$ variables \citep{deFinetti1961bayesian,wang2015general}.
\footnote{
Localization also relates to James-Stein shrinkage;
\citet{efron2010large} connects these dots.
}
Localization decreases the effect of imprecise measurements.
\gls{RPM}s present a broader approach to mitigating mismatch, with
improved performance over localization
(\Cref{sec:empirical,sec:movielens}).

The second theme is robust Bayesian analysis, which studies
sensitivity with respect to the prior \citep{berger1994overview}.
Recent advances directly focus on sensitivity of the posterior
\citep{minsker2014robust,miller2015robust} or the posterior
predictive distribution \citep{kucukelbir2015population}. We draw
connections to these ideas throughout this paper.

The third theme is data reweighting. This involves designing
individual reweighting schemes for specific tasks and models.
Consider robust methods that toss away ``outliers.'' This strategy
involves manually assigning binary weights to datapoints
\citep{huber2011robust}. 
Another example is 
covariate shift
adaptation/importance sampling where reweighting transforms data to
match another target distribution
\citep{veach1995optimally,sugiyama2007covariate,
shimodaira2000improving,wen2014robust}. 
A final example is maximum L$q$-likelihood estimation
\citep{ferrari2010maximum, qin2013maximum, qin2016robust}. Its
solutions can be interpreted as a solution to a weighted likelihood,
whose weights are proportional to a power transformation of the
density. In contrast, \gls{RPM}s treat weights as latent variables.
The weights are automatically inferred; no custom design is required.
\gls{RPM}s also closely connect to ideas around boosting
\citep{schapire2012boosting} and variational tempering
\citep{mandt2016variational}, which also places an exponential weight
on the likelihood term. However, they serve different purposes.
Boosting reweights to build an ensemble of predictors for supervised
learning; variational tempering reweights to escape poor local minima;
\gls{RPM}s reweight to mitigate model mismatch in Bayesian modeling.

\section{Reweighted Probabilistic Models}
\label{sec:weighted}

\glsreset{RPM}
\Glspl{RPM} offer a new approach to robust modeling. The idea is
to automatically identify observations that match the assumptions of
the model and to base posterior inference on these observations.

\subsection{Definitions}

An \gls{RPM} scaffolds over a probabilistic model, $p_\beta(\beta)
\prod_{n=1}^N \ell(y_n\mid\beta)$. Raise each likelihood to a latent
weight and posit a prior on the weights. This gives the reweighted
joint density
\begin{align}
  p(y,\beta,w)
  &=
  \frac{1}{Z}
  p_\beta(\beta)
  p_w(w)
  \prod_{n=1}^N \ell(y_n\mid\beta)^{w_n},
\label{eq:reweighted_joint}
\end{align}
where $Z = \int p_\beta(\beta)
  p_w(w)
  \prod_{n=1}^N \ell(y_n\mid\beta)^{w_n}
  \dif y
  \dif\beta
  \dif w$ is the normalizing factor.

The reweighted density integrates to one when the normalizing factor
$Z$ is finite. This is always true when the likelihood $\ell(\cdot
\mid \beta)$ is an exponential family distribution with Lesbegue base
measure \citep{bernardo2009bayesian}; this is the class of models we
study in this paper.\footnote{Heavy-tailed likelihoods and
Bayesian nonparametric priors may violate this condition; we leave
these for future analysis.}

\begin{figure}[!ht]
\centering
  \begin{subfigure}[b]{1.85in}
    \begin{tikzpicture}
	\begin{pgfonlayer}{nodelayer}
		\node [style=filled] (0) at (0, 0)      {};
		\node [style=hollow] (1) at (-1.5, 0)   {};
		\node [style=empty] (2)  at (-1.5, 0.5) {$\beta$};
		\node [style=empty] (3)  at (0, 0.5)    {$y_n$};
		\node [style=param] (6)  at (-2.5, 0)   {};
		\node [style=empty]  (7) at (-2.5, 0.5) {$p_\beta$};
	\end{pgfonlayer}
	\begin{pgfonlayer}{edgelayer}
		\draw [style=arrow] (1) to (0);
		\draw [style=line]  (6) to (1);
	\end{pgfonlayer}
  \plate {} {(0)(3)} {$N$};
\end{tikzpicture}
    \caption{Original probabilistic model}
    \label{sub:classical_bayes}
  \end{subfigure}

\bigskip
  \begin{subfigure}[b]{2.35in}
    \hspace*{0.2in}
    \begin{tikzpicture}
  \begin{pgfonlayer}{nodelayer}
    \node [style=filled] (0) at (0, 0) {};
    \node [style=hollow] (1) at (-2, 0) {};
    \node [style=empty] (2) at (-2, 0.5) {$\beta$};
    \node [style=empty] (3) at (0, 0.5) {$y_n$};
    \node [style=param] (6) at (-3, 0) {};
    \node [style=empty]  (7) at (-3, 0.5) {$p_\beta$};
    \node [style=hollow] (8) at (-1, -1.0) {};
    \node [style=empty] (9) at (-1, -1.5) {$w_n$};
    \node [style=param] (10) at (-3, -1.0) {};
    \node [style=empty] (11) at (-3, -1.5) {$p_w$};
  \end{pgfonlayer}
  \begin{pgfonlayer}{edgelayer}
    \draw [style=arrow] (1) to (0);
    \draw [style=line]  (6) to (1);
    \draw [style=arrow, bend right]  (8) to (0);
    \draw [style=line]  (10) to (8);
  \end{pgfonlayer}
  \plate {} {(0)(3)(8)} {$N$};
\end{tikzpicture}
    \caption{Reweighted probabilistic model (\textsc{rpm})}
    \label{sub:augmented_bayes}
  \end{subfigure}

\bigskip
  \begin{subfigure}[b]{1.7in}
    \begin{tikzpicture}
  \begin{pgfonlayer}{nodelayer}
    \node [style=filled] (0) at (0, 0)      {};
    \node [style=hollow] (1) at (-1, 0)   {};
    \node [style=empty] (2)  at (-1, 0.5) {$\beta_n$};
    \node [style=empty] (3)  at (0, 0.5)    {$y_n$};
    \node [style=hollow] (6)  at (-2, 0)   {};
    \node [style=empty]  (7) at (-2, 0.5) {$\alpha$};
    \node [style=param] (6)  at (-3, 0)   {};
    \node [style=empty]  (7) at (-3, 0.5) {$p_\alpha$};    
  \end{pgfonlayer}
  \begin{pgfonlayer}{edgelayer}
    \draw [style=arrow] (1) to (0);
    \draw [style=arrow]  (6) to (1);
  \end{pgfonlayer}
  \plate {} {(0)(1)(2)(3)} {$N$};
\end{tikzpicture}
    \caption{Localized probabilistic model}
    \label{sub:localized_bayes}
  \end{subfigure}
  \caption{\gls{RPM}s begin with a probabilistic model \textbf{(a)}
  and introduce a set of weights $w$ as latent variables. This gives
  a model \textbf{(b)} that explores which data observations match
  its assumptions. Localization \textbf{(c)}, instead, builds a
  hierarchical model. (\Cref{app:localization_equiv_RPM} shows when a
  localized model is also an \gls{RPM}.)}
  \label{fig:graphical_models}
\end{figure}
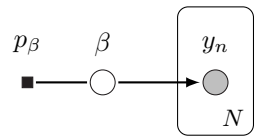
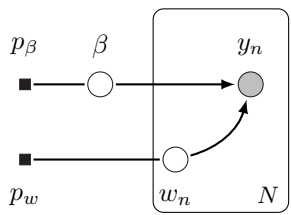
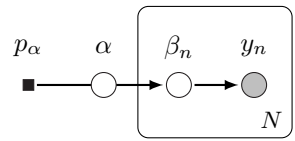

\gls{RPM}s apply to likelihoods that factorize over the
observations. (We discuss non-exchangeable models in
\Cref{sec:discussion}.) \Cref{fig:graphical_models} depicts an
\gls{RPM} as a graphical model. Specific models may have additional
structure, such as a separation of local and global latent variables
\citep{hoffman2013stochastic}, or fixed parameters; we omit these in
this figure.

The reweighted model introduces a set of weights; these are latent
variables, each with support $w_n \in \bbR_{>0}$. To gain intuition,
consider how these weights affect the posterior, which is
proportional to the product of the likelihood of every measurement. A
weight $w_n$ that is close to zero flattens out its corresponding
likelihood $\ell(y_n\mid\beta)^{w_n}$; a weight that is larger than
one makes its likelihood more peaked. This, in turn, enables the
posterior to focus on some measurements more than others. The prior
$p_w(w)$ ensures that not too many likelihood terms get flattened; in
this sense, it plays an important regularization role.

We study three options for this prior on  weights: a bank of Beta
distributions, a scaled Dirichlet distribution, and a bank of Gamma
distributions.

\textbf{Bank of Beta priors.} This option constrains each weight as
$w_n \in (0,1)$. We posit an independent prior for each weight
\begin{align}
  p_w(w) &= \prod_{n=1}^N \text{Beta}(w_n \prm a, b)
\end{align}
and use the same parameters $a$ and $b$ for all weights. This is the
most conservative option for the \gls{RPM}; it ensures that none of
the likelihoods ever becomes more peaked than it was in the original
model.

The parameters $a$, $b$ offer an expressive language to describe
different attitudes towards the weights. For example, setting both
parameters less than one makes the Beta act like a ``two spikes and a
slab'' prior, encouraging weights to be close to zero or one, but not
in between. As another example, setting $a$ greater than $b$
encourages weights to lean towards one.

\textbf{Scaled Dirichlet prior.} This option ensures the sum of the
weights equals $N$. We posit a symmetric Dirichlet prior on all the
weights
\begin{align}
\begin{split}
  w &= Nv\\
  p_v(v) &= \text{Dirichlet}(a\mbone)
\end{split}
\end{align}
where $a$ is a scalar parameter and $\mbone$ is a $(N\times1)$ vector
of ones. In the original model, where all the weights are one, then
the sum of the weights is $N$. The Dirichlet option maintains this
balance; while certain likelihoods may become more peaked, others
will flatten to compensate.

The concentration parameter $a$ gives an intuitive way to configure
the Dirichlet. Small values for $a$ allow the model to easily up- or
down-weight many data observations; larger values for $a$ prefer a
smoother distribution of weights. The Dirichlet option connects to
the bootstrap approaches in \citet{rubin1981bayesian,
kucukelbir2015population}, which also preserves the sum of weights as
$N$.

\textbf{Bank of Gamma priors.} Here we posit an independent Gamma
prior for each weight
\begin{align}
  p_w(w) &= \prod_{n=1}^N \text{Gamma}(w_n\prm a,b)
  \label{eq:prior_gamma}
\end{align}
and use the same parameters $a$ and $b$ for all weights. We do not
recommend this option, because observations can be arbitrarily up- or
down-weighted. In this paper, we only consider \Cref{eq:prior_gamma}
for our theoretical analysis in \Cref{sub:theory}.

The bank of Beta and Dirichlet options perform similarly. We prefer
the Beta option as it is more conservative, yet find the Dirichlet to
be less sensitive to its parameters. We explore these options in the
empirical study (\Cref{sec:empirical}).

\subsection{Theory and intuition}
\label{sub:theory}

How can theory justify Bayesian data reweighting? Here we investigate
its robustness properties. These analyses intend to confirm our
intuition from \Cref{sec:introduction}.
\Cref{app:proof_sketch,app:proof_sketch_2} present proofs in full
technical detail.

\textbf{Intuition}. Recall the logarithm of the \gls{RPM} joint
density from \Cref{eq:logRPM}. Now compute the \gls{MAP} estimate of the weights $w$. The partial
derivative is
\begin{align}
  \frac{\partial \log p(y,\beta,w) }{\partial w_n}
  &=
  \frac{\dif \log p_w(w_n) }{\dif w_n}
  + \log \ell(y_n \mid \beta)
  \label{eq:ddw_log_joint}
\end{align}
for all $n=1,\ldots,N$. Plug the Gamma prior from
\Cref{eq:prior_gamma} into the partial derivative in
\Cref{eq:ddw_log_joint} and set it equal to zero. This gives the
\gls{MAP} estimate of $w_n$,
\begin{align}
  \widehat{w}_n = \frac{a-1}{b - \log \ell(y_n \mid \beta)}.
  \label{eq:ddw_wtfunc}
\end{align}
The \gls{MAP} estimate $\widehat{w}_n$ is an increasing function of
the log likelihood of $y_n$ when $a > 1$.This reveals that $\widehat{w}_n$ shrinks the contribution of observations that are unlikely under the log
likelihood; in turn, this encourages the
\gls{MAP} estimate for $\widehat{\beta}$ to describe the majority of
the observations. This is how an \gls{RPM} makes a probabilistic
model more robust.

A similar argument holds for other exponential family priors on $w$
with $\log w_n$ as a sufficient statistic. We formalize this
intuition and generalize it in the following theorem, which
establishes sufficient conditions where a \gls {RPM} improves the
inference of its latent variables $\beta$.

\begin{theorem}
Denote the true value of $\beta$ as $\beta^*$. Let the posterior mean
of $\beta$ under the weighted and unweighted model be $\bar{\beta}_w$
and $\bar{\beta}_u$ respectively. Assume mild conditions on $p_w$,
$\ell$ and the corruption level, and that
$|\ell(y_n\mid\bar{\beta}_w)-\ell(y_n\mid\beta^*)|<\epsilon$ holds
$\forall n$ with high probability. Then, there exists an $N^*$ such
that for $N>N^*$, we have $\vert\bar{\beta}_u-\beta^*\vert
\succeq_2
\vert\bar{\beta}_w-\beta^*\vert$, where $\succeq_2$ denotes second
order stochastic dominance. (Details in \Cref{app:proof_sketch}.)
\label{theorem:robust}
\end{theorem}

The likelihood bounding assumption is common in robust statistics
theory; it is satisfied for both likely and unlikely (corrupted)
measurements. How much of an improvement does it give? We can
quantify this through the \gls{IF} of $\bar{\beta}_w$.

Consider a distribution $G$ and a statistic $T(G)$ to be a function of
data that comes iid from $G$.  Take a fixed distribution, e.g.,
the population distribution, $F$.  Then, \gls{IF}$(z;T,F)$ measures how
much an additional observation at $z$ affects the statistic
$T(F)$. Define
\begin{align*}
\textsc{if}(z;T,F) &= \lim_{t\rightarrow 0^+}
\frac{T(t\delta_z+(1-t)F)-T(F)} {t}
\end{align*}
for $z$ where this limit exists.  Roughly, the \gls{IF} measures the
asymptotic bias on $T(F)$ caused by a specific observation $z$ that does
not come from $F$.  We consider a statistic $T$ to be robust if its
\gls{IF} is a bounded function of $z$, i.e., if outliers can only exert a limited influence \citep{huber2011robust}.

Here, we study the $\gls{IF}$ of the posterior mean $T =
\bar{\beta}_w$ under the true data generating distribution $F =
\ell(\cdot \mid \beta^*)$. Say a value $z$ has likelihood $\ell(z
\mid \beta^*)$ that is nearly zero; we think of this $z$ as
corrupted. Now consider the weight function induced by the prior
$p_w(w)$. Rewrite it as a function of the log likelihood, like
$w(\log \ell(\cdot \mid \beta^*))$ as in \Cref {eq:ddw_wtfunc}.

\begin{theorem}
If $\lim_{a\rightarrow -\infty}w(a) =0$ and
$\lim_{a\rightarrow -\infty}a\cdot w(a) <\infty$, then
$
\textsc{if}(z;\bar{\beta}_w,
\ell(\cdot\mid\beta^*))\rightarrow 0
\text{ as }\ell(z\mid\beta^*)\rightarrow 0.
$
\label{theorem:influence}
\vspace*{-12pt}
\end{theorem}

This result shows that an \gls{RPM} is robust in that its \gls{IF}
goes to zero for unlikely measurements. This is true for all three
priors. \mbox{(Details in \Cref{app:proof_sketch_2}.)}

\subsection{Inference and computation}

We now turn to inferring the posterior of an \gls{RPM}, $p(\beta, w
\mid y)$. The posterior lacks an analytic closed-form expression for
all but the simplest of models; even if the original model admits
such a posterior for $\beta$, the reweighted posterior may take a
different form.

To approximate the posterior, we appeal to probabilistic programming.
A probabilistic programming system enables a user to write a
probability model as a computer program and then compile that program
into an inference executable. Automated inference is the backbone of
such systems: it takes in a probability model, expressed as a
program, and outputs an efficient algorithm for inference. We use
automated inference in Stan, a probabilistic programming system
\citep{carpenter2015stan}.

In the empirical study that follows, we highlight how \gls{RPM}s
detect and mitigate various forms of model mismatch. As a
common metric, we compare the predictive accuracy on held out data
for the original, localized, and reweighted model.

The posterior predictive likelihood of a new datapoint $y_\dagger$ is
$
p_\text{original}(y_\dagger \mid y)
=
\int
\ell(y_\dagger \mid \beta)
p(\beta \mid y)
\dif \beta.
$
Localization couples each observation with its own copy of the latent
variable; this gives
$
p_\text{localized}(y_\dagger \mid y)
=
\iint \!
\ell(y_\dagger \mid \beta_{\dagger})
p(\beta_{\dagger} \mid \alpha)
p(\alpha \mid y)
\dif\alpha
\dif\beta_{\dagger}
$
where $\beta_{\dagger}$ is the localized latent variable for the new
datapoint. The prior $p(\beta_{\dagger} \mid \alpha)$ has the same
form as $p_\beta$ in \Cref{eq:localization}.

Bayesian data reweighting gives the following posterior predictive
likelihood
\begin{align*}
  p_\textsc{rpm}(y_\dagger \mid y)
  &=
  \iint \!
  p(y_\dagger \mid \beta, w_\dagger)
  p_\textsc{rpm}(\beta \mid y)
  p(w_\dagger)
  \:
  \dif w_\dagger
  \dif \beta,
\end{align*}
where $p_\textsc{rpm}(\beta \mid y)$ is the marginal posterior,
integrating out the inferred weights of the training dataset, and the
prior $p(w_\dagger)$ has the same form as $p_w$ in
\Cref{eq:reweighted_joint}.

\section{Empirical Study}
\label{sec:empirical}

We study \gls{RPM}s under four types of mismatch with reality. This
section involves simulations of realistic scenarios; the next section
presents a recommendation system example using real data. We default
to \gls{NUTS} \citep{hoffman2014nuts} for inference in all
experiments, except for \Cref{sub:gmm,sec:movielens} where we
leverage variational inference \citep{kucukelbir2016automatic}. The
additional computational cost of inferring the weights is
unnoticeable relative to inference in the original model.

\subsection{Outliers: a network wait-time example}
\label{sub:network}

A router receives packets over a network and measures the time it
waits for each packet. Suppose we typically observe wait-times that
follow a Poisson distribution with rate $\beta = 5$. We model each
measurement using a Poisson likelihood $\ell(y_n \mid \beta) =
\text{Poisson}(\beta)$ and posit a Gamma prior on the rate
$p_\beta(\beta) = \Gam (a=2, b=0.5)$.

Imagine that $F$\% percent of the time, the network fails. During
these failures, the wait-times come from a Poisson with much higher
rate $\beta = 50$. Thus, the data actually contains a mixture of two
Poisson distributions; yet, our model only assumes one. (Details in
\Cref{app:gamma_poisson}.)

\begin{figure}[htb]
\centering
   \begin{subfigure}{3.0in}
   \centering
      \input{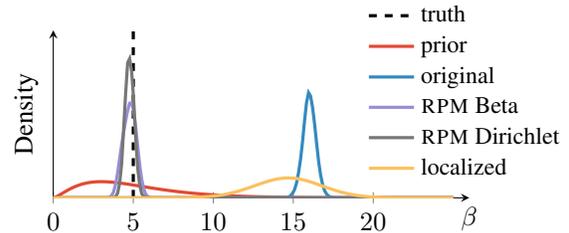}
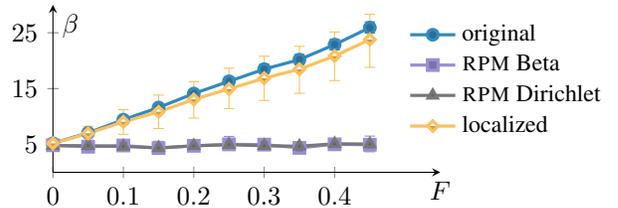
      \vspace*{-10pt}
   \caption{Posteriors for $F = 25\%$ failure rate.}
   \label{fig:poisson45corrupt}
   \end{subfigure}

   \bigskip
   \begin{subfigure}{3.0in}
   \centering
      \begin{tikzpicture}

\definecolor{color1}{rgb}{0.203921568627451,0.541176470588235,0.741176470588235}
\definecolor{color0}{rgb}{0.886274509803922,0.290196078431373,0.2}
\definecolor{color3}{rgb}{0.984313725490196,0.756862745098039,0.368627450980392}
\definecolor{color2}{rgb}{0.596078431372549,0.556862745098039,0.835294117647059}

\begin{axis}[
width=2.65in,
height=1.5in,
xmin=0, xmax=0.55,
ymin=0, ymax=30,
axis x line=bottom,
axis y line=middle,
xlabel={$F$},
x label style={at={(current axis.right of origin)},anchor=north},
ylabel={$\beta$},
xtick={0,0.1,0.2,...,0.4},
ytick={5,15,25},
legend entries={{original}, {\gls{RPM} Beta},{\gls
{RPM} Dirichlet},{localized}},
legend style={draw=none, at={(1.45,0.15)}, anchor=south east, font=\small},
legend cell align=left,
cycle list name=mark list*,
]
\addplot+[
    color1, very thick,
    error bars/.cd,
    y dir=both, y explicit,
]
table[y error plus=ey+, y error minus=ey-]
{
x            y            ey-          ey+
  0.           5.19115091   0.43220633   0.45010775
   0.05         7.07459481   0.50764147   0.53722786
   0.1          9.42234683   0.57274398   0.59839596
   0.15        11.60267583   0.67564777   0.65875432
   0.2         14.10126137   0.72992678   0.72617755
   0.25        16.29750769   0.77952148   0.81409055
   0.3         18.47689903   0.80750852   0.85228689
   0.35        20.18705984   0.87784801   0.92728752
   0.4         22.87417228   0.95609248   0.9271641
   0.45        25.9174598    0.9804041    1.04590511
};

\addplot+[
    color2, very thick,
    error bars/.cd,
    y dir=both, y explicit,
]
table[y error plus=ey+, y error minus=ey-]
{
x            y            ey-          ey+
 0.          4.84646951  0.76291175  0.84843073
  0.05        4.57230001  0.80683137  1.13896967
  0.1         4.80657759  0.94188641  0.63836777
  0.15        4.3343995   0.88730728  1.22753194
  0.2         4.74574944  0.77516682  0.64500185
  0.25        5.00845737  1.08693521  1.41820451
  0.3         4.9599828   0.65283102  1.04392796
  0.35        4.37819242  0.47031712  1.02174061
  0.4         5.01498998  0.9252999   0.96066442
  0.45        4.94253726  1.26434426  1.54347755
};

\addplot+[
    white!46.666666666666664!black, very thick,
    error bars/.cd,
    y dir=both, y explicit,
]
table[y error plus=ey+, y error minus=ey-]
{
x            y            ey-          ey+
 0.          4.77292181  0.60621469  0.64291964
  0.05        4.7716854   0.52025631  0.5283683
  0.1         4.63930071  0.5138655   0.5419149
  0.15        4.37093369  0.55587899  0.63531639
  0.2         4.74705602  0.54851267  0.60661816
  0.25        4.94688575  0.57071228  0.61673264
  0.3         4.75924944  0.60907269  0.63106429
  0.35        4.67317098  0.61492661  0.6730777
  0.4         5.1108797   0.63429037  0.66177764
  0.45        5.05599505  0.59356943  0.65386968
};

\addplot+[
    color3, very thick,
    error bars/.cd,
    y dir=both, y explicit,
]
table[y error plus=ey+, y error minus=ey-]
{
x            y            ey-          ey+
  0.           5.17592944   0.46289187   0.48975379
   0.05         6.96699552   1.17657932   1.2888849
   0.1          9.0372264    2.23650897   2.18518275
   0.15        10.83295063   2.98352064   2.99833936
   0.2         13.05519026   3.33923621   3.17485912
   0.25        14.9237677    3.50603004   3.73580639
   0.3         16.85486087   3.97358882   3.94633098
   0.35        18.45814224   4.30245224   4.11300774
   0.4         20.83520452   4.4023661    4.30728814
   0.45        23.78158842   4.97647858   4.52551246
};

\end{axis}
\end{tikzpicture}
      \vspace*{-10pt}
      \caption{Posterior 95\% credible intervals.}
      \label{fig:poissoncorrupt_CI}
   \end{subfigure}
   \caption{Outliers simulation study. We compare
   $\text{Beta}(0.1, 0.01)$ and $\text{Dir}(\mbone)$ as priors for
   the reweighted probabilistic model. \textbf{(a)} Posterior
   distributions on $\beta$ show a marked difference in detecting the
   correct wait-time rate of $\beta=5$. \textbf{(b)} Posterior 95\%
   confidence intervals across failure rates $F$ show consistent
   behavior for both Beta and Dirichlet priors. ($N=100$ with 50
   replications.)}
   \label{fig:poisson}
\end{figure}

How do we expect an \gls{RPM} to behave in this situation? Suppose
the network failed 25\% of the time. \Cref{fig:poisson45corrupt}
shows the posterior distribution on the rate $\beta$. The original
posterior is centered at 18; this is troubling, not only because the
rate is wrong but also because of how confident the posterior fit is.
Localization introduces greater uncertainty, yet still estimates a
rate around 15. The \gls{RPM} correctly identifies that the majority
of the observations come from $\beta=5$. Observations from when the
network failed are down-weighted. It gives a confident posterior
centered at five.

\Cref{fig:poissoncorrupt_CI} shows posterior 95\% credible intervals
of $\beta$ under failure rates up to $F = 45\%$. The \gls{RPM} is
robust to corrupted measurements; instead it focuses on data that it
can explain within its assumptions. When there is no corruption, the
\gls{RPM} performs just as well as the original model.

\begin{figure}[htbp!]
\centering
\begin{tikzpicture}

\definecolor{color1}{rgb}{0.203921568627451,0.541176470588235,0.741176470588235}

\begin{axis}[
height=1.25in,
width=2.8in,
xlabel={Data Index},
ylabel={Weights},
xmin=0, xmax=100,
ymin=0, ymax=2.1,
tickwidth={0},
xtick={0, 75, 100},
ytick={0,1,2},
ymajorgrids,
bar width=2pt,
bar shift=-1pt,
axis x line*=bottom,
axis y line*=left,
]
\addplot [very thick, gray, dashed]
table {%
75.25 0
75.25 10
};
\addplot [ybar, fill=color1, draw=none]
table {%
1   1.7592860943713
2   1.39481459590098
3   1.67421113937926
4   1.65891367442262
5   1.81279744016377
6   1.63266797632678
7   1.21551010765708
8   1.40168525743965
9   1.19435009905585
10  1.64608000901005
11  1.84342287044214
12  0.830428969018933
13  0.807408783576053
14  0.841777590969254
15  1.19332333070599
16  0.806880553041603
17  0.804864630936155
18  1.35594590698286
19  1.76989159685568
20  0.813204698496986
21  1.20337315814004
22  1.63975159506425
23  1.63918931763792
24  1.81495367911178
25  0.822995578301102
26  0.864047098723105
27  1.66691588549473
28  1.61342266757087
29  1.39962160980893
30  0.847470359553624
31  0.841078234729762
32  1.62452521666575
33  1.37894776554364
34  1.39287431566268
35  0.579004777180068
36  1.40723136820599
37  0.865988697884938
38  1.40638189202231
39  1.6486793568015
40  1.18626051595579
41  0.495773497816017
42  1.63119555116419
43  1.66414590654434
44  0.848837009415355
45  1.65383176713867
46  1.78269332633043
47  0.826845084977367
48  0.425940372999342
49  1.77787640815515
50  1.76354101000462
51  1.85446379631723
52  1.19975039102455
53  1.22515661719285
54  1.20794185334475
55  1.62616095262314
56  1.81199248090837
57  0.833040337960758
58  0.565144232324892
59  1.74017174676113
60  1.21657523739716
61  1.62138006049764
62  1.19350445575359
63  1.80565275257797
64  1.39647414769928
65  0.280172324205834
66  0.591565814764517
67  1.79547241750574
68  1.6568830514181
69  0.870304544567867
70  1.60368331250331
71  1.18038835123895
72  1.78104224091528
73  1.79037958003439
74  1.81765418820537
75  1.80997218692231
76  0.0204417459861507
77  0.0199621382580768
78   0.0183864414954401
79  0.0185333007502802
80  0.0184467295391938
81  0.0170323137813134
82  0.0167210090631106
83  0.0156673776499465
84   0.0156025674732973
85  0.0142012086233576
86   0.014738966723548
87  0.0140325288366903
88  0.0144686414164839
89   0.0134698961081197
90   0.0133410862124034
91   0.0130380831041275
92   0.0131533746454999
93  0.0124981305038118
94  0.0117430235597718
95   0.0110192492014951
96  0.0108410323624218
97  0.0100621895392895
98  0.0101280529931813
99  0.00884469109218876
100 0.00784079705855833
};
\end{axis}

\end{tikzpicture}
\caption{Posterior means of the weights $w$ under the Dirichlet
prior. For visualization purposes, we sorted the data into two
groups: the first 75 contain observations from the normal network;
the remaining 25 are the observations when the network fails.}
\label{fig:poisson25corrupt_wt_dist}
\end{figure}
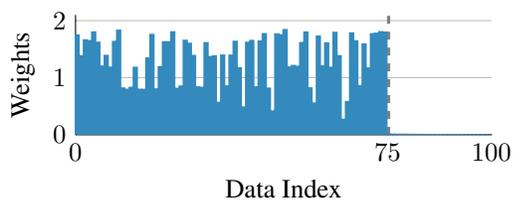

Visualizing the weights elucidates this point.
\Cref{fig:poisson25corrupt_wt_dist} shows the posterior mean
estimates of $w$ for $F = 25\%$. The weights are sorted into two
groups, for ease of viewing. The weights of the corrupted
observations are essentially zero; this downweighting is what allows
the \gls{RPM} to shift its posterior on $\beta$ towards five.

Despite this downweighting, the \gls{RPM} posteriors on $\beta$ are
not overdispersed, as in the localized case. This is due to the
interplay we described in the introduction. Downweighting
observations should lead to a smaller effective sample size, which
would increase posterior uncertainty. But the downweighted datapoints
are corrupted observations; including them also increases posterior
uncertainty.

The \gls{RPM} is insensitive to the prior on the weights; both Beta
and Dirichlet options perform similarly. From here on, we focus on
the Beta option. We let the shape parameter $a$ scale with the data
size $N$ such that $N/a\approx 10^3$; this encodes a mild attitude
towards unit weights. We now move on to other forms of mismatch with
reality.

\subsection{Missing latent groups: predicting color blindness}
\label{sub:glm}

Color blindness is unevenly hereditary: it is much higher for men
than for women \citep{boron2012medical}. Suppose we are not aware of
this fact. We have a dataset of both genders with each individual's
color blindness status and his/her relevant family history. No gender
information is available. Consider analyzing this data using logistic
regression. It can only capture one hereditary group. Thus, logistic
regression misrepresents both groups, even though men exhibit strong
heredity. In contrast, an \gls{RPM} can \emph{detect} and
\emph{mitigate} the missing group effect by focusing on the dominant
hereditary trait. Here we consider men as the dominant group.

We simulate this scenario by drawing binary indicators of color
blindness $y_n \sim \text{Bernoulli}(1/1+\exp(-p_n))$ where the
$p_n$'s come from two latent groups: men exhibit a stronger
dependency on family history ($p_n = 0.5x_n$) than women ($p_n =
0.01x_n$). We simulate family history as $x_n \sim
\text{Unif}(-10,10)$. Consider a Bayesian logistic regression model
without intercept. Posit a prior on the slope as $p_\beta(\beta) =
\cN (0,10)$ and assume a $\text{Beta}(0.1,0.01)$ prior on the
weights. (Details in \Cref{app:glm}.)

\begin{figure}[htb]
\centering
  \begin{tikzpicture}

\definecolor{color1}{rgb}{0.203921568627451,0.541176470588235,0.741176470588235}
\definecolor{color0}{rgb}{0.886274509803922,0.290196078431373,0.2}
\definecolor{color3}{rgb}{0.984313725490196,0.756862745098039,0.368627450980392}
\definecolor{color2}{rgb}{0.596078431372549,0.556862745098039,0.835294117647059}

\begin{axis}[
width=3.0in,
height=1.5in,
xmin=0, xmax=0.55,
ymin=0, ymax=1.50,
axis x line=bottom,
axis y line=middle,
xlabel={$F$},
x label style={at={(current axis.right of origin)},anchor=south},
ylabel={$\beta$},
xtick={0.01,0.11,0.21,0.31,0.41},
xticklabels={0,0.1,0.2,0.3,0.4},
ytick={0,0.5,1},
legend entries={$\beta_\text{men}$, original, \gls{RPM}, localized},
legend style={draw=none, at={(1.2,1.05)}, anchor=north east, font=\small},
legend cell align=left,
cycle list name=mark list*,
]
\addplot [thick, black, opacity=0.5, dashed]
table {%
0 0.5
0.6 0.5
};
\addplot+[
    color1, very thick,
    mark=none,
    error bars/.cd,
    error bar style={line width=1.0pt},
    error mark options={
      rotate=90,
      mark size=4pt,
      line width=1.0pt
    },
    y dir=both, y explicit,
]
table[y error plus=ey+, y error minus=ey-]
{
x            y            ey-          ey+
  0.4         0.19239904  0.05762813  0.06368886
  0.3         0.23343298  0.06443242  0.07070411
  0.2         0.2912805   0.07262494  0.08212068
  0.1         0.38089584  0.08995623  0.10500509
  0.          0.52044044  0.12415286  0.15364387
};

\addplot+[
    color2, very thick,
    mark=none,
    error bars/.cd,
    error bar style={line width=1.0pt},
    error mark options={
      rotate=90,
      mark size=4pt,
      line width=1.0pt
    },
    y dir=both, y explicit,
]
table[y error plus=ey+, y error minus=ey-]
{
x            y            ey-          ey+
  0.42         0.46631579  0.12764987  0.17513351
  0.32         0.54274707  0.1502246   0.21710947
  0.22         0.63673069  0.18192303  0.28077667
  0.12         0.69223216  0.24118793  0.24118793
  0.02          0.80601497  0.33821669  0.33821669
};

\addplot+[
    color3, very thick,
    mark=none,
    error bars/.cd,
    error bar style={line width=1.0pt},
    error mark options={
      rotate=90,
      mark size=4pt,
      line width=1.0pt
    },
    y dir=both, y explicit,
]
table[y error plus=ey+, y error minus=ey-]
{
x            y            ey-          ey+
  0.44        0.7866982   0.50498921  1.68523439
  0.34        0.69793055  0.39296238  1.32271696
  0.24        0.67950786  0.31619449  1.06458282
  0.14        0.6656251   0.26320118  0.79011379
  0.04          0.65101684  0.18601689  0.50790924
};

\end{axis}
\end{tikzpicture}
\caption{Missing latent groups study.
Posterior 95\% credible intervals for the \gls{RPM} always include
the dominant $\beta_\text{men}=0.5$, as we vary the percentage of
females in the data. Dataset size $N=100$ with $50$ replications.}
\label{fig:glm_missgp_CI}
\end{figure}
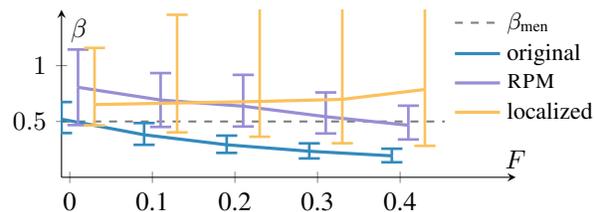

\Cref{fig:glm_missgp_CI} shows the posterior 95\% credible intervals
of $\beta$ as we vary the percentage of females from $F=0\%$ to 40\%.
A horizontal line indicates the correct slope for the dominant group,
$\beta_\text{men} = 0.5$. As the size of the missing latent group
(women) increases, the original model quickly shifts its credible
interval away from $0.5$. The reweighted and localized posteriors
both contain $\beta_\text{men} = 0.5$ for all percentages, but the
localized model exhibits much higher variance in its estimates.

\begin{table*}[tb]
\centering
\begin{tabular}{llccc}
\toprule
\multirow{2}{*}{\textbf{True structure}}
& \multirow{2}{*}{\textbf{Model structure}}
& \textbf{Original}
& \textbf{\gls{RPM}}
& \textbf{Localization}\\
& & mean(std) & mean(std) & mean(std)\\
\midrule
$\beta_0 + \beta_1 x_1 + \beta_2 x_2 + \beta_3 x_1x_2$
& $\beta_0 + \beta_1 x_1 + \beta_2 x_2$
& 3.16(1.37)           & \textbf{2.20}(1.25) & 2.63(1.85)  \\
$\beta_0 + \beta_1 x_1 + \beta_2 x_2 + \beta_3 x_2^2$
& $\beta_0 + \beta_1 x_1 + \beta_2 x_2$
& 30.79(2.60)          & \textbf{16.32}(1.96) & 21.08(5.20)\\
$\beta_0 + \beta_1 x_1 + \beta_2 x_2$
& $\beta_0 + \beta_1 x_1$
& \textbf{0.58}(0.38)  & 0.60(0.40) & 0.98(0.54)  \\
\bottomrule\\
\end{tabular}
\vspace*{-12pt}
\caption{\gls{RPM}s improve absolute deviations of posterior mean
$\beta_1$ estimates. (50 replications.)}
\label{tab:linreg_mismatch}
\end{table*}

This analysis shows how \gls{RPM}s can mitigate the effect of missing
latent groups. While the original logistic regression model would
perform equally poorly on both groups, an \gls{RPM} is able to
automatically focus on the dominant group.

\begin{figure}[!htb]
\centering
\begin{tikzpicture}

\definecolor{color1}{rgb}{0.203921568627451,0.541176470588235,0.741176470588235}
\definecolor{color0}{rgb}{0.886274509803922,0.290196078431373,0.2}
\definecolor{color3}{rgb}{0.984313725490196,0.756862745098039,0.368627450980392}
\definecolor{color2}{rgb}{0.596078431372549,0.556862745098039,0.835294117647059}

\begin{axis}[
width=3.2in,
height=1.25in,
xmin=0, xmax=1.3,
ymin=0, ymax=5,
axis lines=left,
xlabel={$\E_{p(w|y)}[w]$},
x label style={at={(current axis.right of origin)},anchor=north},
ylabel={Density},
ylabel near ticks,
xtick={0,1},
ytick=\empty,
legend entries={{Corrupted ($F=25\%$)},{Clean ($F=0\%$)}},
legend style={draw=none, at={(0.05,0.95)}, anchor=north west,
font=\small},
legend cell align=left,
]
\addplot [ultra thick, color0]
table {%
0 0.0375804712473037
0.0121212121212121 0.05252152430126
0.0242424242424242 0.0715981041960892
0.0363636363636364 0.0952633526693461
0.0484848484848485 0.123799803521855
0.0606060606060606 0.157266210226919
0.0727272727272727 0.195464695854817
0.0848484848484849 0.237935897542008
0.096969696969697 0.283984807098524
0.109090909090909 0.332733677887509
0.121212121212121 0.383192281335487
0.133333333333333 0.434331748720767
0.145454545454545 0.485147637151117
0.157575757575758 0.534701213389433
0.16969696969697 0.582134559079592
0.181818181818182 0.626663166840371
0.193939393939394 0.667556800800783
0.206060606060606 0.704123243404285
0.218181818181818 0.735708736449884
0.23030303030303 0.761723451925792
0.242424242424242 0.781691694165661
0.254545454545455 0.795317319421898
0.266666666666667 0.802547935834575
0.278787878787879 0.803619066044006
0.290909090909091 0.799062505873476
0.303030303030303 0.789670823954299
0.315151515151515 0.776420131487922
0.327272727272727 0.760363015738219
0.339393939393939 0.742510205624369
0.351515151515152 0.723721499817805
0.363636363636364 0.704623599247382
0.375757575757576 0.685566015773374
0.387878787878788 0.666618315576285
0.4 0.647604832450788
0.412121212121212 0.628168255847506
0.424242424242424 0.607851708825368
0.436363636363636 0.586189565374582
0.448484848484848 0.562799101087302
0.460606060606061 0.53746678934296
0.472727272727273 0.51022374362174
0.484848484848485 0.481404333500812
0.496969696969697 0.451681019621861
0.509090909090909 0.422068094273558
0.521212121212121 0.393888391897381
0.533333333333333 0.368700727700932
0.545454545454545 0.348191545579648
0.557575757575758 0.334040831543951
0.56969696969697 0.327778024355742
0.581818181818182 0.330646668249546
0.593939393939394 0.343495752984527
0.606060606060606 0.366710999590487
0.618181818181818 0.400191901879922
0.63030303030303 0.443372164442366
0.642424242424242 0.495274616614786
0.654545454545455 0.554588577352467
0.666666666666667 0.619758713159771
0.678787878787879 0.689078953233426
0.690909090909091 0.760791034693068
0.703030303030303 0.833192149980893
0.715151515151515 0.904757598662367
0.727272727272727 0.974280997688329
0.739393939393939 1.04102670293755
0.751515151515151 1.10487840155461
0.763636363636364 1.16645719289976
0.775757575757576 1.22717509157048
0.787878787878788 1.28918858595999
0.8 1.35522354486051
0.812121212121212 1.42825795099435
0.824242424242424 1.51107171598234
0.836363636363636 1.60570062205145
0.848484848484849 1.71285998832979
0.860606060606061 1.83142724893014
0.872727272727273 1.95808475965777
0.884848484848485 2.08721888761017
0.896969696969697 2.21114524925056
0.909090909090909 2.32068355553909
0.921212121212121 2.40604496669701
0.933333333333333 2.45793128843022
0.945454545454545 2.46869292644762
0.957575757575758 2.43336500086839
0.96969696969697 2.35040774508222
0.981818181818182 2.2220199874782
0.993939393939394 2.053965948143
1.00606060606061 1.85494097382826
1.01818181818182 1.63558241681231
1.03030303030303 1.40728986742077
1.04242424242424 1.18104234907011
1.05454545454545 0.966385639833507
1.06666666666667 0.770716579092043
1.07878787878788 0.598926055529262
1.09090909090909 0.453394377367964
1.1030303030303 0.334276504375505
1.11515151515152 0.239979864807466
1.12727272727273 0.16772731782282
1.13939393939394 0.114109227525123
1.15151515151515 0.0755543948015962
1.16363636363636 0.048681066128949
1.17575757575758 0.0305186927623044
1.18787878787879 0.0186133961258316
1.2 0.011043108528019
};
\addplot [ultra thick, color1, dashed, opacity=0.5]
table {%
0 0.0213489275944965
0.0121212121212121 0.0333196573583606
0.0242424242424242 0.0498429076644722
0.0363636363636364 0.0714751638857802
0.0484848484848485 0.0982778400299359
0.0606060606060606 0.129612774375181
0.0727272727272727 0.164036054422978
0.0848484848484849 0.199357606792352
0.096969696969697 0.232900401998188
0.109090909090909 0.26193559262554
0.121212121212121 0.28420534251806
0.133333333333333 0.298398395846842
0.145454545454545 0.304437372957698
0.157575757575758 0.303480776668597
0.16969696969697 0.297626100389957
0.181818181818182 0.289394744905314
0.193939393939394 0.281149328247312
0.206060606060606 0.274612247782226
0.218181818181818 0.270613937392622
0.23030303030303 0.269116232456793
0.242424242424242 0.269462839740499
0.254545454545455 0.270739905781679
0.266666666666667 0.272108568480681
0.278787878787879 0.273001962968736
0.290909090909091 0.273145565605508
0.303030303030303 0.272434006618211
0.315151515151515 0.270750897437387
0.327272727272727 0.267832216091804
0.339393939393939 0.26324608717245
0.351515151515152 0.256506203170537
0.363636363636364 0.247276422678033
0.375757575757576 0.235584300288337
0.387878787878788 0.221956360477719
0.4 0.207418013538728
0.412121212121212 0.193352129267841
0.424242424242424 0.181260691085762
0.436363636363636 0.172503749776559
0.448484848484848 0.168089464471546
0.460606060606061 0.16856242337585
0.472727272727273 0.173999273187523
0.484848484848485 0.184088151538307
0.496969696969697 0.198252848394333
0.509090909090909 0.215784959963292
0.521212121212121 0.235959653717623
0.533333333333333 0.25812263138956
0.545454545454545 0.281741475244665
0.557575757575758 0.306415399894824
0.56969696969697 0.331840188718197
0.581818181818182 0.35773517147721
0.593939393939394 0.383754990935611
0.606060606060606 0.409421627678335
0.618181818181818 0.43411079754682
0.63030303030303 0.45710694622726
0.642424242424242 0.477709973614582
0.654545454545455 0.495351443207415
0.666666666666667 0.509675314305611
0.678787878787879 0.520563692527721
0.690909090909091 0.528130011756571
0.703030303030303 0.532736097128331
0.715151515151515 0.535092198110151
0.727272727272727 0.536462481869019
0.739393939393939 0.538937646631782
0.751515151515151 0.545683358448949
0.763636363636364 0.561060574595238
0.775757575757576 0.590554847818076
0.787878787878788 0.640529235321212
0.8 0.717885911223931
0.812121212121212 0.829735361224668
0.824242424242424 0.983102210141225
0.836363636363636 1.18456279217483
0.848484848484849 1.43958145537749
0.860606060606061 1.75128666764186
0.872727272727273 2.1185792389243
0.884848484848485 2.53379439042577
0.896969696969697 2.98054525351518
0.909090909090909 3.43267304357302
0.921212121212121 3.85522406427657
0.933333333333333 4.20796201008772
0.945454545454545 4.45117166841898
0.957575757575758 4.55266042619072
0.96969696969697 4.49425407347022
0.981818181818182 4.27599780371634
0.993939393939394 3.91680400883524
1.00606060606061 3.45127577599908
1.01818181818182 2.92352627402459
1.03030303030303 2.3796154214917
1.04242424242424 1.86047016518425
1.05454545454545 1.39680856054311
1.06666666666667 1.00683946818928
1.07878787878788 0.69666711289319
1.09090909090909 0.462680214592213
1.1030303030303 0.294909035891814
1.11515151515152 0.180391846305983
1.12727272727273 0.105887431156126
1.13939393939394 0.0596420782290204
1.15151515151515 0.0322350223945444
1.16363636363636 0.0167170207607428
1.17575757575758 0.00831833462053457
1.18787878787879 0.00397148413805061
1.2 0.00181928820184653
};
\end{axis}

\end{tikzpicture}
  \caption{Kernel density estimate of the distribution of weights across
  all measurements in the missing latent groups study. The percentage
  of females is denoted by $F$. A hypothetical clean dataset receives
  weights that concentrate around one; the actual corrupted dataset
  exhibits a two-hump distribution of weights.}
  \label{fig:weights_corrupted_clean}
\end{figure}
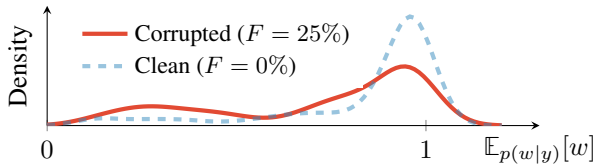

An \gls{RPM} also functions as a diagnostic tool to \emph{detect}
mismatch with reality. The distribution of the inferred weights
indicates the presence of datapoints that defy the assumptions of the
original model. \Cref{fig:weights_corrupted_clean} shows a kernel
density estimate of the inferred posterior weights. A hypothetical
dataset with no corrupted measurements receives weights close to one.
In contrast, the actual dataset with measurements from a missing
latent group exhibit a bimodal distribution of weights. Testing for
bimodality of the inferred weights is one way in which an \gls{RPM}
can be used to diagnose mismatch with reality.

\subsection{Covariate dependence misspecification: a lung cancer risk
study}
\label{sub:misspec}

Consider a study of lung cancer risk. While tobacco usage exhibits a
clear connection, other factors may also contribute. For instance,
obesity and tobacco usage appear to interact, with evidence towards a
quadratic dependence on obesity \citep{odegaard2010bmi}.

Denote tobacco usage as $x_1$ and obesity as $x_2$. We study three
models of lung cancer risk dependency on these covariates. We are
primarily interested in understanding the effect of tobacco usage;
thus we focus on $\beta_1$, the {regression coefficient} for tobacco.
In each model, some form of covariance misspecification discriminates
the true structure from the assumed structure.

For each model, we simulate a dataset of size $N=100$ with random
covariates $x_1\sim \cN(10, 5^2)$ and $x_2\sim \cN(0, 10^2)$ and
regression coefficients $\beta_{0,1,2,3} \sim \text{Unif}(-10,10)$.
Consider a Bayesian linear regression model with prior
$p_\beta(\beta) = \cN(0, 10)$. (Details in \Cref{app:linreg}.)

\Cref{tab:linreg_mismatch} summarizes the misspecification and shows
absolute differences on the estimated ${\beta}_1$ {regression
coefficient}. The \gls{RPM} yields better estimates of $\beta_1$ in
the first two models. These highlight how the \gls{RPM} leverages
datapoints useful for estimating $\beta_1$. The third model is
particularly challenging because obesity is ignored in the
misspecified model. Here, the \gls{RPM} gives similar results to the
original model; this highlights that \gls{RPM}s can only use
available information. Since the original model lacks dependence on
$x_2$, the \gls{RPM} cannot compensate for this.

\begin{table*}[t]
\centering
\begin{tabular}{rSSSSSS}
\toprule
& \multicolumn{2}{c}{\textbf{Outliers}}
& \multicolumn{2}{c}{\textbf{Missing latent groups}}
& \multicolumn{2}{c}{\textbf{Misspecified structure}} \\
{} & \multicolumn{1}{c}{Clean} & \multicolumn{1}{c}{Corrupted}
   & \multicolumn{1}{c}{Clean} & \multicolumn{1}{c}{Corrupted}
   & \multicolumn{1}{c}{Clean} & \multicolumn{1}{c}{Corrupted}  \\
\midrule
{Original model}   & -744.2 & -1244.5
                   & -108.6 & -103.9
                   & -136.3 & -161.7 \\
{Localized model}  & -730.8 & -1258.4
                   & -53.6 & -112.7
                   & -192.5 & -193.1 \\
\gls{RPM}          & \bfseries -328.5 & \bfseries -1146.9
                   & \bfseries -43.9 & \bfseries  -90.5
                   & \bfseries -124.1 & \bfseries -144.1\\
\bottomrule
\end{tabular}
\caption{Posterior predictive likelihoods of clean and corrupted test data.
Outliers and missing latent groups have $F= 25\%$. The misspecified
structure is missing the interaction term. Results are similar for
other levels and types of mismatch with reality.}
\label{tab:asdf}
\end{table*}

\subsection{Predictive likelihood results}

\Cref{tab:asdf} shows how \gls{RPM}s also improve predictive accuracy.
In all the above examples, we simulate test data with and without
their respective types of corruption. \gls{RPM}s improve prediction
for both clean and corrupted data, as they focus on data that match
the assumptions of the original model.

\subsection{Skewed data: cluster selection in a mixture model}
\label{sub:gmm}

Finally, we show how \gls{RPM}s handle skewed data.
The \gls{DPMM} is a versatile model for density estimation and clustering
\citep{bishop2006pattern,murphy2012machine}. While real data may indeed come
from a finite mixture of clusters, there is no reason to assume
each cluster is distributed as a Gaussian.
Inspired by the experiments in
\citet{miller2015robust}, we show how a reweighted \gls{DPMM} reliably recovers
the correct number of components in a mixture of
skewnormals dataset.

\begin{figure}[htb]
\centering
\begin{subfigure}{1.5in}
\centering
   \includegraphics[width=1.5in]{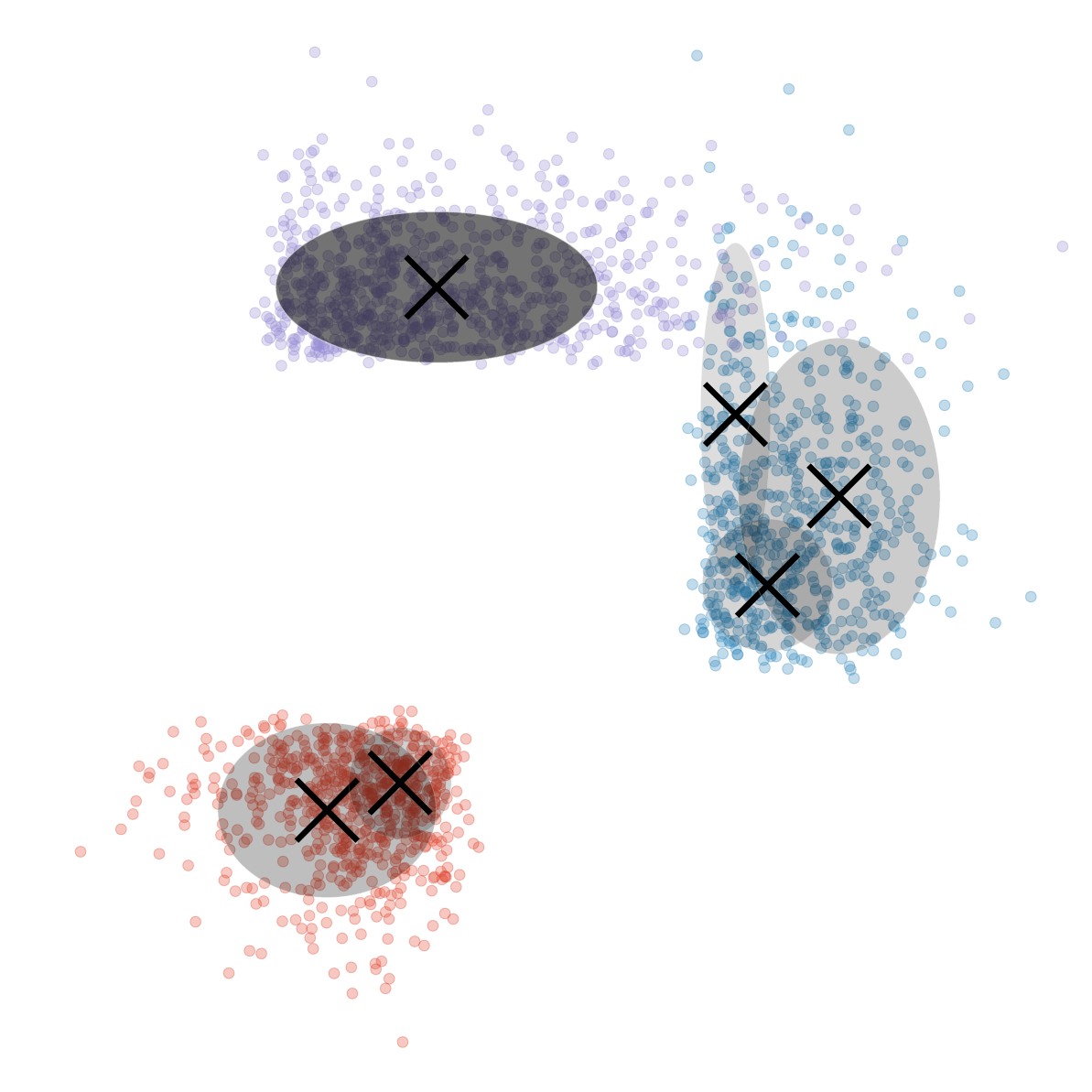}
   \caption{Original model}
   \label{fig:gmm_unwt}
\end{subfigure}
\begin{subfigure}{1.5in}
\centering
   \includegraphics[width=1.5in]{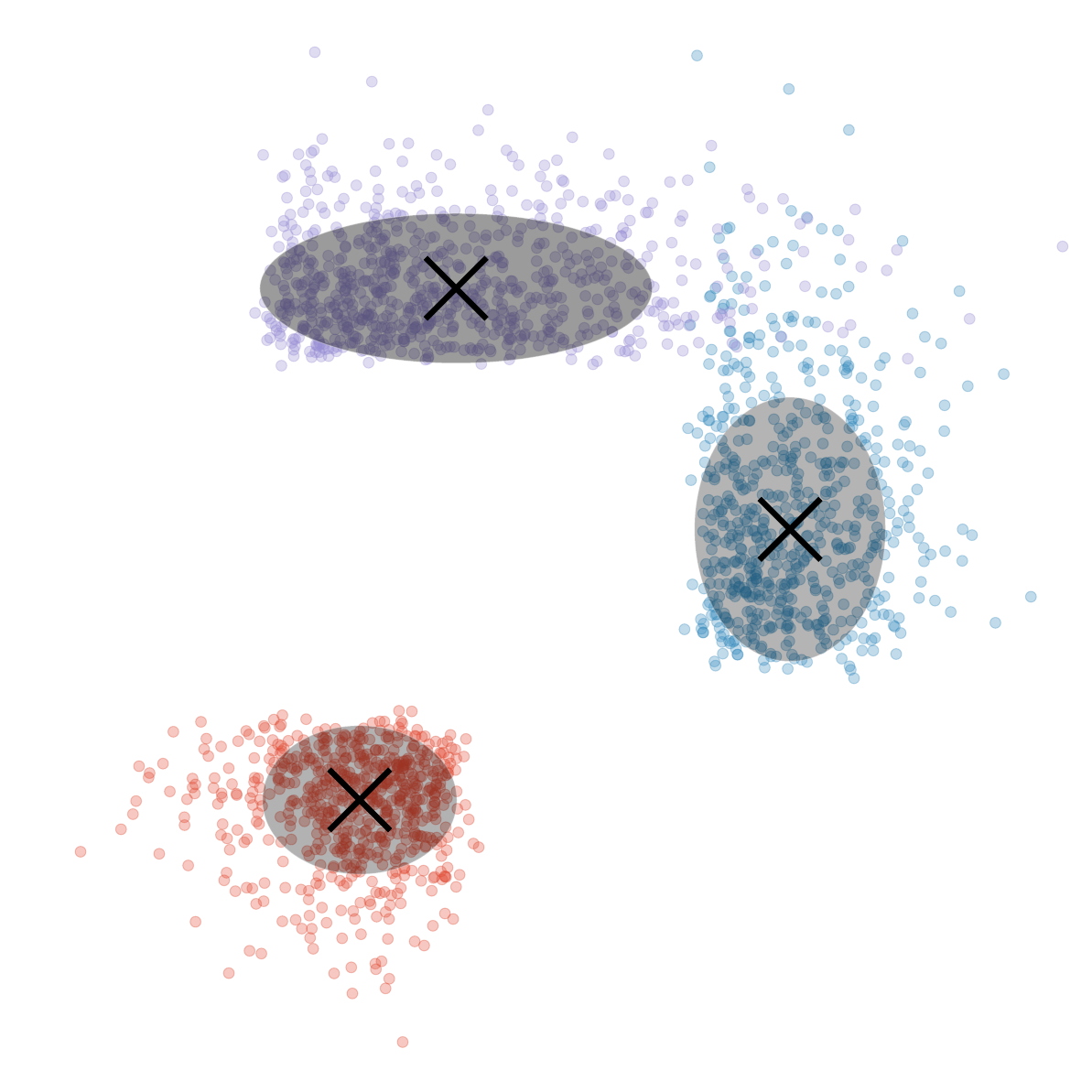}
   \caption{\gls{RPM}}
\label{fig:gmm_wt}
\end{subfigure}
\caption{A finite approximation \gls{DPMM} to skewnormal distributed
data that come from three groups. The shade of each cluster indicates the
inferred mixture proportions ($N=2000$).}
\vspace*{-12pt}
\label{fig:gmm}
\end{figure}

A standard \gls{GMM} with large $K$ and a sparse Dirichlet prior on the mixture
proportions is an approximation to a \gls{DPMM} \citep{ishwaran2012approximate}.
We simulate three clusters from two-dimensional skewnormal
distributions and fit a \gls{GMM} with maximum $K=30$. Here we use \gls{ADVI},
as \gls{NUTS} struggles with inference of mixture models
\citep{kucukelbir2016automatic}. (Details in \Cref{app:skewnormal}.)

\Cref{fig:gmm} shows
posterior mean estimates from the original \gls{GMM}; it incorrectly finds six
clusters. In contrast, the \gls{RPM} identifies the correct three clusters.
Datapoints in the tails of each cluster get down-weighted;
these are datapoints that do not match the Gaussianity assumption of the model.

\glsreset{PF}
\section{Case Study: Poisson factorization for recommendation}
\label{sec:movielens}

We now turn to a study of real data: a recommendation
system. Consider a video streaming service; data comes as a binary
matrix of users and the movies they choose to watch. How
can we identify patterns from such data? \Gls{PF} offers a flexible solution
\citep{cemgil2009bayesian,gopalan2015scalable}. The idea is to infer a $K$-dimensional latent space
of user preferences $\mbtheta$ and movie attributes $\mbbeta$. The inner product
$\mbtheta^\top\mbbeta$ determines the rate of a Poisson
likelihood for each binary measurement; Gamma priors on $\mbtheta$ and $\mbbeta$
promote sparse patterns. As a result, \gls{PF} finds interpretable groupings of
movies, often clustered according to popularity or genre. (Full model in
\Cref{app:PF_model_details}.)

How does classical \gls{PF} compare to its reweighted counterpart? As
input, we use the MovieLens 1M dataset, which contains one million movie ratings
from $6\,000$ users on $4\,000$ movies. We place iid $\text{Gamma}(1,0.001)$
priors on the preferences and attributes. Here, we have the option of
reweighting users or items. We focus on users and place a
$\text{Beta}(100,1)$ prior on their weights. For this model, we use \gls{MAP}
estimation. (Localization is computationally challenging for \gls{PF}; it
requires a separate ``copy'' of $\mbtheta$ for each movie, along with a separate
$\mbbeta$ for each user. This dramatically increases computational cost.)

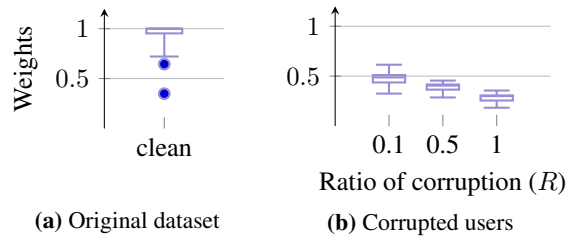
\begin{figure}[htb]
\centering
   \begin{subfigure}{1.5in}
   \centering
    \parbox[c][1.25in]{1.25in}{\vspace*{-15pt}\begin{tikzpicture}

\definecolor{color2}{rgb}{0.596078431372549,0.556862745098039,0.835294117647059}

\begin{axis}[
  width=1.25in,
  height=1.25in,
  boxplot/draw direction=y,
  boxplot/box extend=0.6,
  x axis line style={opacity=0},
  axis x line*=bottom,
  axis y line=left,
  enlarge y limits,
  xmin=-1, xmax=1,
  ymin=0.1, ymax=1.1,
  ylabel={Weights},
  ymajorgrids,
  ytick={0.5,1},
  xtick={0},
  xticklabels={clean}
]
\addplot+[thick, color2, boxplot={draw position=0, whisker range=5.0}]
table[y index=0]{
 0.999988
  0.877964
  0.999985
  0.999991
  0.7592
  0.999936
  0.999977
  0.999983
  0.999989
  0.763275
  0.952671
  0.644639
  0.999977
  0.999989
  0.999988
  0.999989
  0.720892
  0.999991
  0.999895
  0.999985
  0.999989
  0.999991
  0.99998
  0.999987
  0.348545
};

\end{axis}
\end{tikzpicture}}
    \vspace*{-12pt}
    \caption{Original dataset}
    \label{fig:wt_uncorrupt}
\end{subfigure}
\begin{subfigure}{1.5in}
\centering
    \parbox[c][1.25in]{1.75in}{\begin{tikzpicture}

\definecolor{color2}{rgb}{0.596078431372549,0.556862745098039,0.835294117647059}

\begin{axis}[
  width=1.75in,
  height=1.25in,
  boxplot/draw direction=y,
  boxplot/box extend=0.6,
  x axis line style={opacity=0},
  axis x line*=bottom,
  axis y line=left,
  enlarge y limits,
  xmin=0, xmax=4,
  ymin=0.1, ymax=1.1,
  xlabel={Ratio of corruption ($R$)},
  ymajorgrids,
  ytick={0.5,1},
  xtick={1, 2, 3},
  xticklabels={0.1, 0.5, 1}
]

\addplot+[thick, color2, boxplot={draw position=1, whisker range=5.0}]
table[y index=0]{
4.907250000000000223e-01
4.883020000000000693e-01
5.322670000000000456e-01
5.385529999999999484e-01
5.549319999999999808e-01
3.900510000000000366e-01
4.869590000000000307e-01
5.015359999999999818e-01
6.139850000000000030e-01
4.356859999999999622e-01
4.870090000000000252e-01
4.692800000000000304e-01
5.196680000000000188e-01
3.856719999999999593e-01
4.236539999999999750e-01
4.365709999999999869e-01
4.470989999999999687e-01
3.248199999999999976e-01
4.817819999999999880e-01
};

\addplot+[thick, color2, boxplot={draw position=2, whisker range=5.0}]
table[y index=0]{
3.985130000000000061e-01
4.104600000000000470e-01
4.386899999999999689e-01
4.339239999999999764e-01
4.214649999999999785e-01
3.643770000000000064e-01
4.051659999999999706e-01
4.149069999999999703e-01
4.547289999999999943e-01
3.642690000000000095e-01
4.057100000000000151e-01
3.934449999999999892e-01
4.169920000000000293e-01
3.448599999999999999e-01
3.356669999999999932e-01
3.556619999999999782e-01
3.670680000000000054e-01
2.862550000000000372e-01
4.107319999999999860e-01
};

\addplot+[thick, color2, boxplot={draw position=3, whisker range=5.0}]
table[y index=0]{
3.000110000000000277e-01
3.020229999999999859e-01
3.388510000000000133e-01
3.361870000000000136e-01
3.028760000000000341e-01
2.690719999999999779e-01
3.004129999999999856e-01
3.021619999999999862e-01
3.555449999999999999e-01
2.560080000000000133e-01
2.998870000000000147e-01
3.096690000000000276e-01
2.982630000000000003e-01
2.501599999999999935e-01
2.401300000000000101e-01
2.516849999999999921e-01
2.548949999999999827e-01
1.832380000000000120e-01
3.263690000000000202e-01
};

\end{axis}
\end{tikzpicture}}
    \vspace*{-12pt}
    \caption{Corrupted users}
    \label{fig:wt_corrupt}
    \end{subfigure}
\caption{Inferred weights for clean and corrupted data. \textbf{(a)} Most
users receive weights very close to one. \textbf{(b)} Corrupted users receive
weights much smaller than one. Larger ratios of corruption $R$ imply lower
weights.}
\end{figure}

\begin{table}[bt]
\centering
\begin{tabular}{rccc}
\toprule
\multirow{2}{*}{\textbf{Average log likelihood}}
&
\multicolumn{3}{c}{\textbf{Corrupted users}}\\
& \multicolumn{1}{c}{0\%} & \multicolumn{1}{c}{1\%} & \multicolumn{1}{c}{2\%} \\
\midrule
{Original model}   &  $-1.68$ & $-1.73$ & $-1.74$\\
\gls{RPM} &  \mb{-1.53} & \mb{-1.53} & \mb{-1.52}\\
\bottomrule

\end{tabular}
\caption{Held-out predictive accuracy under varying amounts of corruption.
Held-out users chosen randomly ($20\%$ of total users).}
\label{tab:movie_pred}
\end{table}

We begin by analyzing the original (clean) dataset. Reweighting improves the
average
held-out log likelihood from $-1.68$ of the original model to $-1.53$ of the
corresponding
\gls{RPM}. The boxplot in
\Cref{fig:wt_uncorrupt}
shows the inferred weights. The majority of
users receive weight one, but a few users are down-weighted. These are film
enthusiasts who appear to indiscriminately watch many movies from many genres.
(\Cref{app:list_of_downweighted_movies} shows an example.) These users do
not contribute towards identifying movies that go together; this explains why
the \gls{RPM} down-weights them.

Recall the example from our introduction. A child typically watches popular
animated films, but her parents occasionally use her account to watch horror
films. We simulate this by corrupting a small percentage of users. We replace
a ratio $R = (0.1, 0.5, 1)$ of these users' movies with randomly selected
movies.

The boxplot in \Cref{fig:wt_corrupt} shows the weights we infer
for these corrupted users, based on how many of their movies we randomly
replace. The weights decrease as we corrupt more movies. \Cref{tab:movie_pred}
shows how this leads to higher held-out predictive accuracy; down-weighting
these corrupted users leads to better prediction.

\glsreset{RPM}
\section{Discussion}
\label{sec:discussion}

\Glspl{RPM} offer a systematic approach to mitigating various forms
of mismatch with reality. The idea is to raise each data likelihood
to a weight and to infer the weights along with the hidden patterns.
We demonstrate how this strategy introduces robustness and improves
prediction accuracy across four types of mismatch.

\gls{RPM}s also offer a way to \emph{detect} mismatch with reality.
The distribution of the inferred weights sheds light onto datapoints
that fail to match the original model's assumptions. \gls{RPM}s can
thus lead to new model development and deeper insights about our
data.

\gls{RPM}s can also work with non-exchangeable data, such as time
series. Some time series models admit exchangeable likelihood
approximations \citep{guinness2013transformation}. For other models,
a non-overlapping windowing approach would also work. The idea of
reweighting could also extend to structured likelihoods, such as
Hawkes process models.

\section*{Acknowledgements} 
We thank Adji Dieng, Yuanjun Gao, Inchi Hu, Christian Naesseth,
Rajesh Ranganath, Francisco Ruiz, Dustin Tran, and Joshua Vogelstein for their
insightful pointers and comments. This work is supported by NSF IIS-1247664, ONR
N00014-11-1-0651, DARPA PPAML FA8750-14-2-0009, DARPA SIMPLEX
N66001-15-C-4032, and the Alfred P. Sloan Foundation.
\putbib[BIB1]
\end{bibunit}

\clearpage
\begin{bibunit}[apa]
{\onecolumn
\appendix

\section{Localized generalized linear model as an \gls{RPM}}
\label{app:localization_equiv_RPM}

Localization in \glspl{GLM} is equivalent to reweighting, with constraints on
the weight function $w(\cdot)$ induced by $p_w$. We prelude the theorem with a simple illustration in linear regression.

Consider $N$ iid observations $\{(x_n, y_n)\}^N_1$. We regress $y$ against $x$: \[y_n = \beta_1 (x_n-\bar{x}) + \beta_0 + \epsilon_n, \epsilon_n\stackrel{iid}{\sim} N(0, \sigma^2),\]
where $\bar{x} = \sum^N_{n=1}x_n$. The maximum likelihood estimate of $(\beta_0, \beta_1)$ is
\[(\widehat{\beta}_0, \widehat{\beta}_1) = \text{argmin}_{\beta_0,\beta_1}\sum^N_{n=1}(y_n-\beta_1 (x_n-\bar{x})-\beta_0)^2.\]
The localized model is
\[y_n = \beta_{1n}\times (x_n-\bar{x}) + \beta_{0} + \epsilon_n, \beta_{1n}\stackrel{iid}{\sim} N(\beta_1, \lambda^2), \epsilon_n\stackrel{iid}{\sim} N(0, \sigma^2),\]
where $\{\beta_{1n}\}^N_{n=1} \independent \{\epsilon_n\}^N_{n=1}$. Marginalizing out $\beta_{1n}$'s gives
\[y_n = \beta_1\times (x_n-\bar{x}) + \beta_0+ \gamma_n, \gamma_n\stackrel{iid}{\sim} N(0, (x_n-\bar{x})^2\cdot \lambda^2+\sigma^2).\]
The maximum likelihood estimate of $(\beta_0, \beta_1)$ in the localized model thus becomes
\[(\widehat{\beta}_0, \widehat{\beta}_1) = \text{argmin}_{\beta_0,\beta_1}\sum^N_{n=1}\frac{(y_n-\beta_1(x_n-\bar{x})-\beta_0)^2}{(x_n-\bar{x})^2\cdot \lambda^2+\sigma^2}.\]
This is equivalent to the reweighting approach with
\[w_n = \frac{1}{(x_n-\bar{x})^2\cdot \lambda^2+\sigma^2}.\]

We generalize this argument into generalized linear models.\\

\begin{theorem}
Localization in a \gls{GLM} with identity link infers $\beta_1$ from
\begin{align*}
y_n \mid x_n, \beta_{1n}, \beta_0
&\sim
\exp\left(\frac{y_n\cdot\eta_n-b_1(\eta_n)}{a_1(\phi)}+c_1(y_n,\phi)\right),\\
\eta_n
&=
\beta_0+\beta_{1n} \cdot(x_n-\bar{x}),\\
\beta_{1n}\mid\beta_1
&\sim
\exp\left(\frac{\beta_{1n}\cdot\beta_1-b_2(\beta_1)}{a_2(\nu)}+c_2(\beta_
{1n},\nu)\right),
\end{align*}
where $a_1(\cdot), a_2(\cdot)$ denote dispersion constants, $b_1(\cdot), b_2(\cdot)$ denote normalizing constants, and $c_1(\cdot), c_2(\cdot)$ denote carrier densities of exponential family distributions.

Inferring $\beta_1$ from this localized \gls{GLM} is equivalent to inferring $\beta_1$ from the reweighted model with weights
\begin{align*}
w_n
&=
\E_{p(\beta_{1n}|\beta_1)}
\left[
\exp\left(
\frac{(y_n-E(y_n\mid \beta_0+\tilde{\beta}_{1n}(x_n-\bar
{x})))(\beta_{1n}-\beta_1) (x_n-\bar{x})}{a_1(\phi)}
\right)
\right]
\end{align*}
for some $\{\tilde{\beta}_{1n}\}^N_1$.
\end{theorem}

\textbf{Proof} A classical \gls{GLM} with an identity link is
\[y_n\sim \exp\left(\frac{y_n\cdot\eta_n-b_1(\eta_n)}{a_1(\phi)}+c_1
(y_n,\phi)\right),\]
\[\eta_n = \beta_0+\beta_1 \cdot(x_n-\bar{x}),\]
whose maximum likelihood estimate calculates
\[(\widehat{\beta}_0, \widehat{\beta}_1) = \text{argmax}_{\beta_0,\beta_1}\prod^N_{n=1}L_{c,n},\]
where
\[L_{c,n}=\exp\left(\frac{y_n\cdot(\beta_0+\beta_1 (x_n-\bar{x}))-b_1(\beta_0+\beta_1(x_n-\bar{x}))}{a_1(\phi)}+c_1(y_n,\phi)\right).\]
On the other hand, the maximum likelihood estimate of the localized model calculates
\[(\widehat{\beta}_0, \widehat{\beta}_1) = \text{argmax}_{\beta_0,\beta_1}\prod^N_{n=1}L_{l,n},\]
where
\[\begin{aligned}
L_{l,n}= \int\exp\left(\frac{y_n\cdot(\beta_0+\beta_{1n} (x_n-\bar{x}))-b_1(\beta_0+\beta_{1n} (x_n-\bar{x}))}{a_1(\phi)}+c_1(y_n,\phi)\right.\\
\left. +\frac{\beta_{1n}\beta_1-b_2(\beta_1)}{a_2(\nu)}+c_2(\beta_{1n},\nu)\right)d\beta_{1n}.
\end{aligned}\]
A localized \gls{GLM} is thus reweighting the likelihood term of each observation by
\begin{align*}
\frac{L_{l,n}}{L_{c,n}}=&\int\exp\left(\frac{y_n(\beta_{1n}-\beta_1) (x_n-\bar{x})-b_1(\beta_0+\beta_{1n}(x_n-\bar{x}))+b_1(\beta_0+\beta_{1} (x_n-\bar{x}))}{a_1(\phi)}\right.\\
&\left. +\frac{\beta_{1n}\beta_1-b_2(\beta_1)}{a_2(\nu)}+c_2(\beta_{1n},\nu)\right)d\beta_{1n}\\
=&\int\exp\left(\frac{y_n(\beta_{1n}-\beta_1) (x_n-\bar{x})-b_1'(\beta_0+\tilde{\beta}_{1n}(x_n-\bar{x}))(\beta_{1n}-\beta_1) (x_n-\bar{x})}{a_1(\phi)}\right.\\
&\left.+\frac{\beta_{1n}\beta_1-b_2(\beta_1)}{a_2(\nu)}+c_2(\beta_{1n},\nu)\right)d\beta_{1n}\\
=&\int\exp\left(\frac{(y_n-b_1'(\beta_0+\tilde{\beta}_{1n}(x_n-\bar{x})))(\beta_{1n}-\beta_1) (x_n-\bar{x})}{a_1(\phi)}+\frac{\beta_{1n}\beta_1-b_2(\beta_1)}{a_2(\nu)}\right.\\
&\left.+c_2(\beta_{1n},\nu)\right)d\beta_{1n}\\
=&E_{p(\beta_{1n}|\beta_1)}\exp\left(\frac{(y_n-E(y_n\mid \beta_0+\tilde{\beta}_{1n}(x_n-\bar{x})))(\beta_{1n}-\beta_1) (x_n-\bar{x})}{a_1(\phi)}\right)
\end{align*}
where $\tilde{\beta}_{1n}$ is some value between $\beta_1$ and $\beta_{1n}$ and the second equality is due to mean value theorem. The last equality is due to $y_n$ residing in the exponential family.

\hspace{\fill}$\blacksquare$

\clearpage
\section{Proof sketch of \cref{theorem:robust}}
\label{app:proof_sketch}
Denote as $\ell(y\mid\beta:\beta\in\Theta)$ the statistical model we fit to the data set $y_1, ..., y_N\stackrel{iid}{\sim}\bar{P}_N$. $\ell(\cdot|\beta)$ is a density function with respect to some carrier measure $\nu(dy),$ and $\Theta$ is the parameter space of $\beta$.

Denote the desired true value of $\beta$ as
$\beta_0$. Let $p_0(d\beta)$ be the prior measure absolute continuous in a neighborhood of $\beta_0$ with a continuous density at $\beta_0$. Let $p_w(dw)$ be the prior measure on weights $(w_n)^N_{n=1}$. Finally, let the posterior mean of $\beta$ under the weighted and unweighted
model be $\bar{\beta}_w$ and $\bar{\beta}_u$ and the corresponding \gls{MLE} be
$\widehat{\beta}_w$ and $\widehat{\beta}_u$ respectively.

Let us start with some assumptions.

\begin{assumption}
\label{assumption:first}
$\ell(\cdot|\beta)$ is twice-differentiable and log-concave.
\end{assumption}

\begin{assumption}
There exist an increasing function $w(\cdot):\mathbb{R}\rightarrow\mathbb{R}^+$ such that $w_n = w(\log\ell(y_n|\beta))$ solves
\[\frac{\partial}{\partial w_n}p_w((w_n)^N_{n=1})+\log\ell(y_n|\beta)=0, n = 1, ..., N.\]
\end{assumption}
We can immediately see that the bank of $\text{Beta}(\alpha, \beta)$ priors with $\alpha>1$ and the bank of $\text{Gamma}(k, \theta)$ priors with $k>1$ satisfy this condition.

\begin{assumption}
\label{assumption:likeclose2truth}
$P(|\log\ell(y_n\mid\widehat{\beta}_w)-\log\ell(y_n\mid\beta_0)|<\epsilon)>1-\delta_1$ holds
$\forall n$ for some $\epsilon, \delta_1 > 0$.
\end{assumption}

This assumption includes the following two cases: (1) $\widehat{\beta}_w$ is
close to the true parameter $\beta_0$, i.e. the corruption is not at all
influential in parameter estimation, and (2) deviant points in $y_1, ..., y_N$ are far enough from typical observations coming from $\ell(y\mid\beta_0)$ that $\log\ell(y_n\mid\widehat{\beta}_w)$ and $\log\ell(y_n\mid\beta_0)$ almost coincide. This assumption precisely explains why the \gls{RPM} performs well in \Cref{sec:empirical}.

\begin{assumption}
\label{assumption:bound}
$|\widehat{\beta}_u-\beta_0|\geq M$ for some $M$.
\end{assumption}

\begin{assumption}
\label{assumption:last}
There exist a permutation $\pi(i):\{1, ..., N\}\rightarrow \{1, ..., N\}$ s.t. \[P(\sum^k_{n=1}\frac{\log\ell(y_{\pi(i)}|\beta_0)'}{\sum^N_{n=1}\log\ell(y_{\pi(i)}|\beta_0)'}\leq(1-\frac{4\epsilon}{M})\sum^k_{n=1}\frac{\log\ell(y_{\pi(i)}|\tilde{\beta}_n)''}{\sum^N_{n=1}\log\ell(y_{\pi(i)}|\check{\beta}_n)''}, k = 1, ..., n-1)\geq 1-\delta_2,\]
for $\tilde{\beta}_n$ and $\check{\beta}_n$ between $\widehat{\beta}_u$ and $\beta_0$ and for some $\delta_2>0$.
\end{assumption}
By noticing that $\sum^N_{n=1}\frac{\log\ell(y_n|\beta_0)'}{\sum^N_{n=1}\log\ell(y_n|\beta_0)'} = 1$, $\sum^N_{n=1}\frac{\log\ell(y_n|\tilde{\beta}_n)''}{\sum^N_{n=1}\log\ell(y_n|\check{\beta}_n)''}(1-\frac{4\epsilon}{M})\approx 1$, and $\text{Var}(\log\ell(y_n|\beta)')>>\text{Var}(\log\ell(y_n|\beta)'')$ in general,

this assumption is not particularly restrictive. For instance, a normal likelihood has $\text{Var}(\log\ell(y_n|\beta)'') = 0$.

\textbf{Theorem}
Assume \Cref{assumption:first}-\Cref{assumption:last}. There exists an $N^*$ such that for $N>N^*$, we have
$\vert\bar{\beta}_u-\beta_0\vert
\succeq_2
\vert\bar{\beta}_w-\beta_0\vert$, where $\succeq_2$ denotes second order stochastic dominance.

\emph{Proof Sketch.} We resort to \gls{MAP} estimates of $\{w_n\}_1^N$ and
$\delta_1=\delta_2=0$ for simplicity of the sketch.

By Bernstein-von Mises theorem, there exists $N^*$ s.t. $N>N^*$ implies the
posterior means $\bar{\beta}_w$ and $\bar{\beta}_u$ are close to their
corresponding \glspl{MLE} $\widehat{\beta}_w$ and $\widehat{\beta}_u$. Thus it
is sufficient to show instead that $\vert\widehat{\beta}_u-\beta_0\vert (1-\frac{4\epsilon}{M})
\succeq_2
(\vert\widehat{\beta}_w-\beta_0\vert)$.

By mean value theorem, we have
\[|\hat{\beta}_w-\beta_0| = \frac{-\sum^N_{n=1}w(\log\ell(y_n|\beta_0))(\log\ell(y_n|\beta_0)')}{\sum^N_{n=1}w(\log\ell(y_n|\beta_0))(\log\ell(y_n|\tilde{\beta}_n)'')}\]
and
\[|\hat{\beta}_u-\beta_0| = \frac{-\sum^N_{n=1}\log\ell(y_n|\beta_0)'}{\sum^N_{n=1}\log\ell(y_n|\check{\beta}_n)''},\]
where $\tilde{\beta}_n$ and $\check{\beta}_n$ are between $\hat{\beta}_u$ and $\beta_0$.

It is thus sufficient to show
\[|\sum^N_{n=1}w(\log\ell(y_n|\beta_0))\frac{\log\ell(y_n|\tilde{\beta}_n)''}{\sum^N_{n=1}\log\ell(y_n|\check{\beta}_n)''}(1-\frac{4\epsilon}{M})| \succeq_2
|\sum^N_{n=1}w(\log\ell(y_n|\beta_0))\frac{\log\ell(y_n|\beta_0)'}{\sum^N_{n=1}\log\ell(y_n|\beta_0)'}|\]
This is true by \Cref{assumption:last} and a version of stochastic majorization inequality (e.g. Theorem 7 of \citet{egozcue2010gains}). \hfill$\blacksquare$

The whole proof of Theorem \ref{theorem:robust} is to formalize the intuitive argument that if we downweight an observation whenever it deviates from the truth of $\beta_0$, our posterior estimate will be closer to $\beta_0$ than without downweighting, given the presence of these disruptive observations.

\clearpage
\section{Proof sketch of \cref{theorem:influence}}
\label{app:proof_sketch_2}

We again resort to \gls{MAP} estimates of weights for simplicity. Denote a probability distribution with a $t$-mass at $z$ as $P_t = t\delta_z+(1-t)P_{\beta_0}$. By differentiating the estimating equation
\[\int \{w(\log\ell(z\mid\beta))\log\ell'(z\mid\beta)\}P_t(z)dz=0\]
with respect to $t$, we obtain that
\[\gls{IF}(z;\widehat{\beta}_w, \ell(\cdot\vert\beta_0)) =
J_w(\beta_0)^{-1}\{w(\log\ell(z\mid\beta_0))\log \ell^\prime(z\vert\beta_0)\},\]
where
\[J_w(\beta_0) = \E_{\ell(z\vert\beta_0)}
\left[w(\log\ell(z\mid\beta_0))
\log \ell^\prime(z\vert\beta_0)
\log \ell^\prime(z\vert\beta_0)^\top\right].\]

It is natural to consider $z$ with $\log\ell(z\mid\beta_0)$ negatively large as an outlier. By investigating the behavior of $w(a)$ as $a$ goes to $-\infty$, we can easily see that
\begin{align*}
\textsc{if}(z;\widehat{\beta}_w,
\ell(\cdot\mid\beta_0))\rightarrow 0,
\text{ as }\ell(z\mid\beta_0)\rightarrow 0,
\end{align*}
if \[\lim_{a\rightarrow -\infty}w(a) =0\text{ and }
\lim_{a\rightarrow -\infty}a\cdot w(a) <\infty.\]

\clearpage
\section{Empirical study details}
\label{app:empirical_study_details}

We present details of the four models in \Cref{sec:empirical}.

\subsection{Corrupted observations}
\label{app:gamma_poisson}
We generate a data set $\{y_n\}_1^N$ of size $N=100$, $(1-F)\cdot N$ of them from Poisson(5) and $F\cdot N$ of them from Poisson(50). The corruption rate $F$ takes values from 0, 0.05, 0.10, ..., 0.45.

The localized Poisson model is
\[\{y_n\}_1^N\mid \{\theta_n\}^N_1 \sim \prod^N_{n=1}\text{Poisson}(y_n\mid\theta_n),\]
\[\theta_n\mid\theta \stackrel{iid}{\sim} \cN(\theta,\sigma^2),\]
with priors
\[\theta\sim \text{Gamma}(\gamma_a,\gamma_b),\]
\[\sigma^2\sim\text{lognormal}(0, \nu^2).\]

The \gls{RPM} is
\[p(\{y_n\}_1^N\mid \theta, \{w_n\}_1^N) = \left[\prod^N_{n=1}\text{Poisson}
(y_n;\theta)^{w_n}\right]\text{Gamma}(\theta|2, 0.5)\left[\prod^N_{n=1}\text
{Beta}(w_n;0.1,
0.01)\right].\]

\subsection{Missing latent groups}
\label{app:glm}

We generate a data set $\{(y_n, x_n\}_1^N$ of size $N=100$; $x_n \sim
\text{Unif}(-10,10)$; $y_n \sim \text{Bernoulli}(1/1+\exp(-p_n))$ where $(1-F)\cdot N$ of them from $p_n = 0.5x_n$ and $F\cdot N$ of them from $p_n = 0.01x_n$. The missing latent group size $F$ takes values from 0, 0.05, 0.10, ..., 0.45.

The localized model is
\[y\mid x \sim \prod^N_{n=1}\text{Bernoulli}(y_n\mid\text{logit}(\beta_{1n}x_n)),\]
\[\beta_{1n}\sim\cN(\beta_1,\sigma^2),\]
with priors
\[\beta_1\sim\cN (0, \tau^2),\]
\[\sigma^2\sim \text{Gamma}(\gamma_a, \gamma_b).\]

The \gls{RPM} is
\begin{align*}
p(\{y_n\}_1^N, \beta, \{w_n\}_1^N\mid \{x_n\}_1^N) = &\left[\prod^N_{n=1}\text{Bernoulli}(y_n;1/1+\exp(-\beta x_n))^{w_n}\right]\cN(\beta;0, 10)\\
\times &\left[\prod^N_{n=1}\text{Beta}(w_n;0.1, 0.01)\right].
\end{align*}

\subsection{Covariate dependence misspecification}
\label{app:linreg}

We generate a data set $\{(y_n, x_{1n}, x_{2n})\}_1^N$ of size $N=100$; $x_{1n}\stackrel{iid}{\sim}\cN(10, 5^2)$, $x_{2n}\stackrel{iid}{\sim}\cN(0, 10^2)$, $\beta_{0,1,2,3} \stackrel{iid}{\sim} \text{Unif}(-10,10)$, $\epsilon_n\stackrel{iid}{\sim} \cN(0,1).$

\begin{enumerate}
\item Missing an interaction term

Data generated from $y_n = \beta_0 + \beta_1 x_1 + \beta_2 x_2 + \beta_3 x_1x_2 + \epsilon_n$.

The localized model is
\[y\mid (x_1, x_2)\sim \prod^N_{n=1}\cN(y_n\mid\beta_{0n}+\beta_{1n}x_{1n}+\beta_{2n}x_{2n},\sigma^2),\]
\[\beta_{jn}\mid\beta_j\stackrel{iid}{\sim}\cN(\beta_j,\sigma_j^2), \]
with priors
\[\beta_j\stackrel{iid}{\sim}\cN (0, \tau^2), j = 0, 1, 2,\]
\[\sigma_j^2\stackrel{iid}{\sim}\text{lognormal}(0, \nu^2), j = 0, 1, 2,\]
\[\sigma^2\sim \text{Gamma}(\gamma_a, \gamma_b).\]

The \gls{RPM} is
\begin{align*}
p\left(
\{y_n\}_1^N, \beta_{0,1,2}, \{w_n\}_1^N \mid \{x_{1n}, x_{2n}\}_1^N)
\right)
&=
\left[
\prod^N_{n=1}\cN(y_n;\beta_0 + \beta_1 x_1 + \beta_2 x_2, \sigma^2)^
{w_n}
\right]\\
&\quad\times
\text{Gamma}(\sigma^2; 1, 1)\\
&\quad\times
\prod^2_{j=0}\cN(\beta_j; 0, 10)
\left[\prod^N_{n=1}\text{Beta}(w_n;0.1, 0.01)\right].
\end{align*}

\item Missing a quadratic term

Data generated from $y_n = \beta_0 + \beta_1 x_1 + \beta_2 x_2 + \beta_3 x_2^2 + \epsilon_n$.

The localized model is
\[y\mid (x_1, x_2)\sim \prod^N_{n=1}\cN(y_n\mid\beta_{0n}+\beta_{1n}x_{1n}+\beta_{2n}x_{2n},\sigma^2),\]
\[\beta_{jn}\mid\beta_j\stackrel{iid}{\sim}\cN(\beta_j,\sigma_j^2), \]
with priors
\[\beta_j\stackrel{iid}{\sim}\cN (0, \tau^2), j = 0, 1, 2,\]
\[\sigma_j^2\stackrel{iid}{\sim}\text{lognormal}(0, \nu^2), j = 0, 1, 2,\]
\[\sigma^2\sim \text{Gamma}(\gamma_a, \gamma_b).\]

The \gls{RPM} is
\begin{align*}
p\left(
\{y_n\}_1^N, \beta_{0,1,2}, \{w_n\}_1^N \mid \{x_{1n}, x_{2n}\}_1^N)
\right)
&=
\left[
\prod^N_{n=1}\cN(y_n;\beta_0 + \beta_1 x_1 + \beta_2 x_2, \sigma^2)^
{w_n}
\right]\\
&\quad\times
\text{Gamma}(\sigma^2; 1, 1)\\
&\quad\times
\prod^2_{j=0}\cN(\beta_j; 0, 10)
\left[\prod^N_{n=1}\text{Beta}(w_n;0.1, 0.01)\right].
\end{align*}

\item Missing a covariate

Data generated from $y_n = \beta_0 + \beta_1 x_1 + \beta_2 x_2 + \epsilon_n$.

The localized model is
\[y\mid (x_1)\sim \prod^N_{n=1}\cN(y_n\mid\beta_{0n}+\beta_{1n}x_{1n},\sigma^2),\]
\[\beta_{jn}\mid\beta_j\stackrel{iid}{\sim}\cN(\beta_j,\sigma_j^2), \]
with priors
\[\beta_j\stackrel{iid}{\sim}\cN (0, \tau^2), j = 0, 1,\]
\[\sigma_j^2\stackrel{iid}{\sim}\text{lognormal}(0, \nu^2), j = 0, 1,\]
\[\sigma^2\sim \text{Gamma}(\gamma_a, \gamma_b).\]

The \gls{RPM} is
\begin{align*}
p\left(
\{y_n\}_1^N, \beta_{0,1}, \{w_n\}_1^N \mid \{x_{1n}\}_1^N)
\right)
&=
\left[
\prod^N_{n=1}\cN(y_n;\beta_0 + \beta_1 x_1, \sigma^2)^
{w_n}
\right]\\
&\quad\times
\text{Gamma}(\sigma^2; 1, 1)\\
&\quad\times
\prod^1_{j=0}\cN(\beta_j; 0, 10)
\left[\prod^N_{n=1}\text{Beta}(w_n;0.1, 0.01)\right].
\end{align*}

\end{enumerate}

\subsection{Skewed distributions}
\label{app:skewnormal}

We generate a data set $\{(x_{1n}, x_{2n})\}_1^N$ of size $N=2000$ from a mixture of three skewed normal distributions, with location parameters (-2, -2), (3, 0), (-5, 7), scale parameters (2, 2), (2, 4), (4, 2), shape parameters -5, 10, 15, and mixture proportions 0.3, 0.3, 0.4. So the true number of components in this data set is 3.

The \gls{RPM} is
\begin{align*}
&p(\{(x_{1n}, x_{2n})\}_1^N, \{\mu_k\}_1^{30}, \{\Sigma_k\}_1^{30}, \{\pi_k\}_1^{30}, \{w_n\}_1^N ) \\= &\left[\prod^N_{n=1}[\sum_{k=1}^{30}\pi_k\cN((x_{1n}, x_{2n};\mu_k, \Sigma_k)]^{w_n}\right]\left[\prod^{30}_{k=1}\cN(\mu_{k,1}; 0, 10)\cN(\mu_{k,2}; 0, 10)\right]\\&\times\left[\prod^{30}_{k=1}\text{lognormal}(\sigma_{k,1}; 0, 10)\text{lognormal}(\sigma_{k,2}; 0, 10)\right]\\&\times\text{Dirichlet}((\pi_k)_1^{30};\mathbf{1})\left[\prod^N_{n=1}\text{Beta}(w_n; 1, 0.05)\right],
\end{align*}
where $\mu_k = (\mu_{k,1}, \mu_{k,2})$ and $\Sigma_k = \left( \begin{array}{cc}
\sigma^2_{k,1} & 0 \\
0 & \sigma^2_{k,2} \\
\end{array} \right)$.

\clearpage
\section{Poisson factorization model}
\label{app:PF_model_details}
Poisson factorization models a matrix of count data as a low-dimensional inner
product \citep{cemgil2009bayesian,gopalan2015scalable}.

Consider a data set of a matrix sized $U\times I$ with non-negative integer elements $x_{ui}$. In the recommendation example, we have $U$ users and $I$ items and each $x_{ui}$ entry being the rating of user $u$ on item $i$.

The user-reweighted \gls{RPM} is
\begin{align*}
p(\{x_{ui}\}_{U\times I}, \{\theta_u\}_1^U, \{\beta_i\}_1^I) =& \left[\prod_{u=1}^U[\prod_{i=1}^I\text{Poisson}(x_{ui}; \theta_u\top\beta_i)]^{w_u}\right]\\
&\times\left[\prod_{u=1}^U\prod_{k=1}^K\text{Gamma}(\theta_{u,k};1, 0.001)\right]\left[\prod_{i=1}^I\prod_{k=1}^K\text{Gamma}(\beta_{i,k};1, 0.001)\right]\\
&\times\prod^U_{u=1}\text{Beta}(w_u;100, 1),
\end{align*}
where $K$ is the number of latent dimensions.

\parhead{Dataset}. We use the Movielens-1M data set: user-movie ratings collected from a movie recommendation service.\footnote{\url{http://grouplens.org/datasets/movielens/}}

\clearpage
\section{Profile of a downweighted user}
\label{app:list_of_downweighted_movies}

Here we show a donweighted user in the \gls{RPM} analysis of the Movielens
1M dataset. This user watched $325$ movies; we rank her movies according to
their popularity in the dataset.

%
\begin{center}
\small
\begin{supertabular}{rrllr}
\toprule
 \textbf{Title} & \textbf{Genres} & $\%$ \\ 
\midrule
Usual Suspects, The (1995) & Crime$|$Thriller & 45.0489 \\ 
  2001: A Space Odyssey (1968) & Drama$|$Mystery$|$Sci-Fi$|$Thriller & 41.6259 \\ 
  Ghost (1990) & Comedy$|$Romance$|$Thriller & 32.0293 \\ 
  Lion King, The (1994) & Animation$|$Children's$|$Musical & 30.7457 \\ 
  Leaving Las Vegas (1995) & Drama$|$Romance & 27.3533 \\ 
  Star Trek: Generations (1994) & Action$|$Adventure$|$Sci-Fi & 27.0171 \\ 
  African Queen, The (1951) & Action$|$Adventure$|$Romance$|$War & 26.1614 \\ 
  GoldenEye (1995) & Action$|$Adventure$|$Thriller & 25.1222 \\ 
  Birdcage, The (1996) & Comedy & 19.7433 \\ 
  Much Ado About Nothing (1993) & Comedy$|$Romance & 18.6125 \\ 
  Hudsucker Proxy, The (1994) & Comedy$|$Romance & 17.1760 \\ 
  My Fair Lady (1964) & Musical$|$Romance & 17.1760 \\ 
  Philadelphia Story, The (1940) & Comedy$|$Romance & 15.5562 \\ 
  James and the Giant Peach (1996) & Animation$|$Children's$|$Musical & 13.8142 \\ 
  Crumb (1994) & Documentary & 13.1724 \\ 
  Remains of the Day, The (1993) & Drama & 12.9279 \\ 
  Adventures of Priscilla, Queen of the Desert, The (1994) & Comedy$|$Drama & 12.8362 \\ 
  Reality Bites (1994) & Comedy$|$Drama & 12.4389 \\ 
  Notorious (1946) & Film-Noir$|$Romance$|$Thriller & 12.0416 \\ 
  Brady Bunch Movie, The (1995) & Comedy & 11.9499 \\ 
  Roman Holiday (1953) & Comedy$|$Romance & 11.8888 \\ 
  Apartment, The (1960) & Comedy$|$Drama & 11.6748 \\ 
  Rising Sun (1993) & Action$|$Drama$|$Mystery & 11.1858 \\ 
  Bringing Up Baby (1938) & Comedy & 11.1553 \\ 
  Bridges of Madison County, The (1995) & Drama$|$Romance & 10.9413 \\ 
  Pocahontas (1995) & Animation$|$Children's$|$Musical & 10.8802 \\ 
  Hunchback of Notre Dame, The (1996) & Animation$|$Children's$|$Musical & 10.8191 \\ 
  Mr. Smith Goes to Washington (1939) & Drama & 10.6663 \\ 
  His Girl Friday (1940) & Comedy & 10.5134 \\ 
  Tank Girl (1995) & Action$|$Comedy$|$Musical$|$Sci-Fi & 10.4218 \\ 
  Adventures of Robin Hood, The (1938) & Action$|$Adventure & 10.0856 \\ 
  Eat Drink Man Woman (1994) & Comedy$|$Drama & 9.9939 \\ 
  American in Paris, An (1951) & Musical$|$Romance & 9.7188 \\ 
  Secret Garden, The (1993) & Children's$|$Drama & 9.3215 \\ 
  Short Cuts (1993) & Drama & 9.0465 \\ 
  Six Degrees of Separation (1993) & Drama & 8.8325 \\ 
  First Wives Club, The (1996) & Comedy & 8.6797 \\ 
  Age of Innocence, The (1993) & Drama & 8.3435 \\ 
  Father of the Bride (1950) & Comedy & 8.2213 \\ 
  My Favorite Year (1982) & Comedy & 8.1601 \\ 
  Shadowlands (1993) & Drama$|$Romance & 8.1601 \\ 
  Some Folks Call It a Sling Blade (1993) & Drama$|$Thriller & 8.0990 \\ 
  Little Women (1994) & Drama & 8.0379 \\ 
  Kids in the Hall: Brain Candy (1996) & Comedy & 7.9768 \\ 
  Cat on a Hot Tin Roof (1958) & Drama & 7.7017 \\ 
  Corrina, Corrina (1994) & Comedy$|$Drama$|$Romance & 7.3961 \\ 
  Muppet Treasure Island (1996) & Adventure$|$Comedy$|$Musical & 7.3655 \\ 
  39 Steps, The (1935) & Thriller & 7.2127 \\ 
  Farewell My Concubine (1993) & Drama$|$Romance & 7.2127 \\ 
  Renaissance Man (1994) & Comedy$|$Drama$|$War & 7.1210 \\ 
  With Honors (1994) & Comedy$|$Drama & 6.7543 \\ 
  Virtuosity (1995) & Sci-Fi$|$Thriller & 6.7543 \\ 
  Cold Comfort Farm (1995) & Comedy & 6.4792 \\ 
  Man Without a Face, The (1993) & Drama & 6.4181 \\ 
  East of Eden (1955) & Drama & 6.2958 \\ 
  Three Colors: White (1994) & Drama & 5.9597 \\ 
  Shadow, The (1994) & Action & 5.9291 \\ 
  Boomerang (1992) & Comedy$|$Romance & 5.6846 \\ 
  Hellraiser: Bloodline (1996) & Action$|$Horror$|$Sci-Fi & 5.6540 \\ 
  Basketball Diaries, The (1995) & Drama & 5.5318 \\ 
  My Man Godfrey (1936) & Comedy & 5.3790 \\ 
  Very Brady Sequel, A (1996) & Comedy & 5.3484 \\ 
  Screamers (1995) & Sci-Fi$|$Thriller & 5.2567 \\ 
  Richie Rich (1994) & Children's$|$Comedy & 5.1956 \\ 
  Beautiful Girls (1996) & Drama & 5.1650 \\ 
  Meet Me in St. Louis (1944) & Musical & 5.1650 \\ 
  Ghost and Mrs. Muir, The (1947) & Drama$|$Romance & 4.9817 \\ 
  Waiting to Exhale (1995) & Comedy$|$Drama & 4.9817 \\ 
  Boxing Helena (1993) & Mystery$|$Romance$|$Thriller & 4.7983 \\ 
  Belle de jour (1967) & Drama & 4.7983 \\ 
  Goofy Movie, A (1995) & Animation$|$Children's$|$Comedy & 4.6760 \\ 
  Spitfire Grill, The (1996) & Drama & 4.6760 \\ 
  Village of the Damned (1995) & Horror$|$Sci-Fi & 4.6149 \\ 
  Dracula: Dead and Loving It (1995) & Comedy$|$Horror & 4.5232 \\ 
  Twelfth Night (1996) & Comedy$|$Drama$|$Romance & 4.5232 \\ 
  Dead Man (1995) & Western & 4.4927 \\ 
  Miracle on 34th Street (1994) & Drama & 4.4621 \\ 
  Halloween: The Curse of Michael Myers (1995) & Horror$|$Thriller & 4.4315 \\ 
  Once Were Warriors (1994) & Crime$|$Drama & 4.3704 \\ 
  Kid in King Arthur's Court, A (1995) & Adventure$|$Comedy$|$Fantasy & 4.3399 \\ 
  Road to Wellville, The (1994) & Comedy & 4.3399 \\ 
  Restoration (1995) & Drama & 4.2176 \\ 
  Oliver \& Company (1988) & Animation$|$Children's & 4.0648 \\ 
  Basquiat (1996) & Drama & 3.9731 \\ 
  Pagemaster, The (1994) & Adventure$|$Animation$|$Fantasy & 3.8814 \\ 
  Giant (1956) & Drama & 3.8509 \\ 
  Surviving the Game (1994) & Action$|$Adventure$|$Thriller & 3.8509 \\ 
  City Hall (1996) & Drama$|$Thriller & 3.8509 \\ 
  Herbie Rides Again (1974) & Adventure$|$Children's$|$Comedy & 3.7897 \\ 
  Backbeat (1993) & Drama$|$Musical & 3.6675 \\ 
  Umbrellas of Cherbourg, The (1964) & Drama$|$Musical & 3.5758 \\ 
  Ruby in Paradise (1993) & Drama & 3.5452 \\ 
  Mrs. Winterbourne (1996) & Comedy$|$Romance & 3.4841 \\ 
  Bed of Roses (1996) & Drama$|$Romance & 3.4841 \\ 
  Chungking Express (1994) & Drama$|$Mystery$|$Romance & 3.3619 \\ 
  Free Willy 2: The Adventure Home (1995) & Adventure$|$Children's$|$Drama & 3.3313 \\ 
  Party Girl (1995) & Comedy & 3.2702 \\ 
  Solo (1996) & Action$|$Sci-Fi$|$Thriller & 3.1785 \\ 
  Stealing Beauty (1996) & Drama & 3.1479 \\ 
  Burnt By the Sun (Utomlyonnye solntsem) (1994) & Drama & 3.1479 \\ 
  Naked (1993) & Drama & 2.9034 \\ 
  Kicking and Screaming (1995) & Comedy$|$Drama & 2.9034 \\ 
  Jeffrey (1995) & Comedy & 2.8729 \\ 
  Made in America (1993) & Comedy & 2.8423 \\ 
  Lawnmower Man 2: Beyond Cyberspace (1996) & Sci-Fi$|$Thriller & 2.8117 \\ 
  Davy Crockett, King of the Wild Frontier (1955) & Western & 2.7812 \\ 
  Vampire in Brooklyn (1995) & Comedy$|$Romance & 2.7506 \\ 
  NeverEnding Story III, The (1994) & Adventure$|$Children's$|$Fantasy & 2.6895 \\ 
  Candyman: Farewell to the Flesh (1995) & Horror & 2.6284 \\ 
  Air Up There, The (1994) & Comedy & 2.6284 \\ 
  High School High (1996) & Comedy & 2.5978 \\ 
  Young Poisoner's Handbook, The (1995) & Crime & 2.5367 \\ 
  Jane Eyre (1996) & Drama$|$Romance & 2.5367 \\ 
  Jury Duty (1995) & Comedy & 2.4756 \\ 
  Girl 6 (1996) & Comedy & 2.4450 \\ 
  Farinelli: il castrato (1994) & Drama$|$Musical & 2.3227 \\ 
  Chamber, The (1996) & Drama & 2.2616 \\ 
  Blue in the Face (1995) & Comedy & 2.2005 \\ 
  Little Buddha (1993) & Drama & 2.2005 \\ 
  King of the Hill (1993) & Drama & 2.1699 \\ 
  Shanghai Triad (Yao a yao yao dao waipo qiao) (1995) & Drama & 2.1699 \\ 
  Scarlet Letter, The (1995) & Drama & 2.1699 \\ 
  Blue Chips (1994) & Drama & 2.1394 \\ 
  House of the Spirits, The (1993) & Drama$|$Romance & 2.1394 \\ 
  Tom and Huck (1995) & Adventure$|$Children's & 2.0477 \\ 
  Life with Mikey (1993) & Comedy & 2.0477 \\ 
  For Love or Money (1993) & Comedy & 2.0171 \\ 
  Princess Caraboo (1994) & Drama & 1.9560 \\ 
  Addiction, The (1995) & Horror & 1.9560 \\ 
  Mrs. Parker and the Vicious Circle (1994) & Drama & 1.9254 \\ 
  Cops and Robbersons (1994) & Comedy & 1.9254 \\ 
  Wonderful, Horrible Life of Leni Riefenstahl, The (1993) & Documentary & 1.8949 \\ 
  Strawberry and Chocolate (Fresa y chocolate) (1993) & Drama & 1.8949 \\ 
  Bread and Chocolate (Pane e cioccolata) (1973) & Drama & 1.8643 \\ 
  Of Human Bondage (1934) & Drama & 1.8643 \\ 
  To Live (Huozhe) (1994) & Drama & 1.8337 \\ 
  Now and Then (1995) & Drama & 1.8337 \\ 
  Flipper (1996) & Adventure$|$Children's & 1.8032 \\ 
  Mr. Wrong (1996) & Comedy & 1.8032 \\ 
  Before and After (1996) & Drama$|$Mystery & 1.7115 \\ 
  Maya Lin: A Strong Clear Vision (1994) & Documentary & 1.6504 \\ 
  Horseman on the Roof, The (Hussard sur le toit, Le) (1995) & Drama & 1.6504 \\ 
  Moonlight and Valentino (1995) & Drama$|$Romance & 1.6504 \\ 
  Andre (1994) & Adventure$|$Children's & 1.6504 \\ 
  House Arrest (1996) & Comedy & 1.6198 \\ 
  Celtic Pride (1996) & Comedy & 1.6198 \\ 
  Amateur (1994) & Crime$|$Drama$|$Thriller & 1.6198 \\ 
  White Man's Burden (1995) & Drama & 1.5892 \\ 
  Heidi Fleiss: Hollywood Madam (1995) & Documentary & 1.5892 \\ 
  Adventures of Pinocchio, The (1996) & Adventure$|$Children's & 1.5892 \\ 
  National Lampoon's Senior Trip (1995) & Comedy & 1.5587 \\ 
  Angel and the Badman (1947) & Western & 1.5587 \\ 
  Poison Ivy II (1995) & Thriller & 1.5281 \\ 
  Bitter Moon (1992) & Drama & 1.4976 \\ 
  Perez Family, The (1995) & Comedy$|$Romance & 1.4670 \\ 
  Georgia (1995) & Drama & 1.4364 \\ 
  Love in the Afternoon (1957) & Comedy$|$Romance & 1.4059 \\ 
  Inkwell, The (1994) & Comedy$|$Drama & 1.4059 \\ 
  Bloodsport 2 (1995) & Action & 1.4059 \\ 
  Bad Company (1995) & Action & 1.3753 \\ 
  Underneath, The (1995) & Mystery$|$Thriller & 1.3753 \\ 
  Widows' Peak (1994) & Drama & 1.3447 \\ 
  Alaska (1996) & Adventure$|$Children's & 1.2836 \\ 
  Jefferson in Paris (1995) & Drama & 1.2531 \\ 
  Penny Serenade (1941) & Drama$|$Romance & 1.2531 \\ 
  Big Green, The (1995) & Children's$|$Comedy & 1.2531 \\ 
  What Happened Was... (1994) & Comedy$|$Drama$|$Romance & 1.2531 \\ 
  Great Day in Harlem, A (1994) & Documentary & 1.1919 \\ 
  Underground (1995) & War & 1.1919 \\ 
  House Party 3 (1994) & Comedy & 1.1614 \\ 
  Roommates (1995) & Comedy$|$Drama & 1.1614 \\ 
  Getting Even with Dad (1994) & Comedy & 1.1308 \\ 
  Cry, the Beloved Country (1995) & Drama & 1.1308 \\ 
  Stalingrad (1993) & War & 1.1308 \\ 
  Endless Summer 2, The (1994) & Documentary & 1.1308 \\ 
  Browning Version, The (1994) & Drama & 1.1308 \\ 
  Fluke (1995) & Children's$|$Drama & 1.1002 \\ 
  Scarlet Letter, The (1926) & Drama & 1.1002 \\ 
  Pyromaniac's Love Story, A (1995) & Comedy$|$Romance & 1.0697 \\ 
  Castle Freak (1995) & Horror & 1.0697 \\ 
  Double Happiness (1994) & Drama & 1.0697 \\ 
  Month by the Lake, A (1995) & Comedy$|$Drama & 1.0391 \\ 
  Once Upon a Time... When We Were Colored (1995) & Drama & 1.0391 \\ 
  Favor, The (1994) & Comedy$|$Romance & 1.0086 \\ 
  Manny \& Lo (1996) & Drama & 1.0086 \\ 
  Visitors, The (Les Visiteurs) (1993) & Comedy$|$Sci-Fi & 1.0086 \\ 
  Carpool (1996) & Comedy$|$Crime & 0.9780 \\ 
  Total Eclipse (1995) & Drama$|$Romance & 0.9780 \\ 
  Panther (1995) & Drama & 0.9474 \\ 
  Lassie (1994) & Adventure$|$Children's & 0.9474 \\ 
  It's My Party (1995) & Drama & 0.9169 \\ 
  Kaspar Hauser (1993) & Drama & 0.9169 \\ 
  It Takes Two (1995) & Comedy & 0.9169 \\ 
  Purple Noon (1960) & Crime$|$Thriller & 0.8863 \\ 
  Nadja (1994) & Drama & 0.8557 \\ 
  Haunted World of Edward D. Wood Jr., The (1995) & Documentary & 0.8557 \\ 
  Dear Diary (Caro Diario) (1994) & Comedy$|$Drama & 0.8252 \\ 
  Faces (1968) & Drama & 0.8252 \\ 
  Love \& Human Remains (1993) & Comedy & 0.7946 \\ 
  Man of the House (1995) & Comedy & 0.7946 \\ 
  Curdled (1996) & Crime & 0.7641 \\ 
  Jack and Sarah (1995) & Romance & 0.7641 \\ 
  Denise Calls Up (1995) & Comedy & 0.7641 \\ 
  Aparajito (1956) & Drama & 0.7641 \\ 
  Hunted, The (1995) & Action & 0.7641 \\ 
  Colonel Chabert, Le (1994) & Drama$|$Romance$|$War & 0.7335 \\ 
  Thin Line Between Love and Hate, A (1996) & Comedy & 0.7335 \\ 
  Nina Takes a Lover (1994) & Comedy$|$Romance & 0.7335 \\ 
  Ciao, Professore! (Io speriamo che me la cavo ) (1993) & Drama & 0.7029 \\ 
  In the Bleak Midwinter (1995) & Comedy & 0.7029 \\ 
  Naked in New York (1994) & Comedy$|$Romance & 0.7029 \\ 
  Maybe, Maybe Not (Bewegte Mann, Der) (1994) & Comedy & 0.6724 \\ 
  Police Story 4: Project S (Chao ji ji hua) (1993) & Action & 0.6418 \\ 
  Algiers (1938) & Drama$|$Romance & 0.6418 \\ 
  Tom \& Viv (1994) & Drama & 0.6418 \\ 
  Cold Fever (A koldum klaka) (1994) & Comedy$|$Drama & 0.6112 \\ 
  Amazing Panda Adventure, The (1995) & Adventure$|$Children's & 0.6112 \\ 
  Marlene Dietrich: Shadow and Light (1996) & Documentary & 0.6112 \\ 
  Jupiter's Wife (1994) & Documentary & 0.6112 \\ 
  Stars Fell on Henrietta, The (1995) & Drama & 0.6112 \\ 
  Careful (1992) & Comedy & 0.5807 \\ 
  Kika (1993) & Drama & 0.5807 \\ 
  Loaded (1994) & Drama$|$Thriller & 0.5501 \\ 
  Killer (Bulletproof Heart) (1994) & Thriller & 0.5501 \\ 
  Clean Slate (Coup de Torchon) (1981) & Crime & 0.5501 \\ 
  Killer: A Journal of Murder (1995) & Crime$|$Drama & 0.5501 \\ 
  301, 302 (1995) & Mystery & 0.5196 \\ 
  New Jersey Drive (1995) & Crime$|$Drama & 0.5196 \\ 
  Gold Diggers: The Secret of Bear Mountain (1995) & Adventure$|$Children's & 0.4890 \\ 
  Spirits of the Dead (Tre Passi nel Delirio) (1968) & Horror & 0.4890 \\ 
  Fear, The (1995) & Horror & 0.4890 \\ 
  From the Journals of Jean Seberg (1995) & Documentary & 0.4890 \\ 
  Celestial Clockwork (1994) & Comedy & 0.4584 \\ 
  They Made Me a Criminal (1939) & Crime$|$Drama & 0.4584 \\ 
  Man of the Year (1995) & Documentary & 0.4584 \\ 
  New Age, The (1994) & Drama & 0.4279 \\ 
  Reluctant Debutante, The (1958) & Comedy$|$Drama & 0.4279 \\ 
  Savage Nights (Nuits fauves, Les) (1992) & Drama & 0.4279 \\ 
  Faithful (1996) & Comedy & 0.4279 \\ 
  Land and Freedom (Tierra y libertad) (1995) & War & 0.4279 \\ 
  Boys (1996) & Drama & 0.3973 \\ 
  Big Squeeze, The (1996) & Comedy$|$Drama & 0.3973 \\ 
  Gumby: The Movie (1995) & Animation$|$Children's & 0.3973 \\ 
  All Things Fair (1996) & Drama & 0.3973 \\ 
  Kim (1950) & Children's$|$Drama & 0.3667 \\ 
  Infinity (1996) & Drama & 0.3667 \\ 
  Peanuts - Die Bank zahlt alles (1996) & Comedy & 0.3667 \\ 
  Ed's Next Move (1996) & Comedy & 0.3667 \\ 
  Hour of the Pig, The (1993) & Drama$|$Mystery & 0.3667 \\ 
  Walk in the Sun, A (1945) & Drama & 0.3667 \\ 
  Death in the Garden (Mort en ce jardin, La) (1956) & Drama & 0.3362 \\ 
  Collectionneuse, La (1967) & Drama & 0.3362 \\ 
  They Bite (1996) & Drama & 0.3362 \\ 
  Original Gangstas (1996) & Crime & 0.3362 \\ 
  Gordy (1995) & Comedy & 0.3362 \\ 
  Last Klezmer, The (1995) & Documentary & 0.3056 \\ 
  Butterfly Kiss (1995) & Thriller & 0.3056 \\ 
  Talk of Angels (1998) & Drama & 0.3056 \\ 
  In the Line of Duty 2 (1987) & Action & 0.3056 \\ 
  Tarantella (1995) & Drama & 0.3056 \\ 
  Under the Domin Tree (Etz Hadomim Tafus) (1994) & Drama & 0.2751 \\ 
  Dingo (1992) & Drama & 0.2751 \\ 
  Billy's Holiday (1995) & Drama & 0.2751 \\ 
  Venice/Venice (1992) & Drama & 0.2751 \\ 
  Low Life, The (1994) & Drama & 0.2751 \\ 
  Phat Beach (1996) & Comedy & 0.2751 \\ 
  Catwalk (1995) & Documentary & 0.2751 \\ 
  Fall Time (1995) & Drama & 0.2445 \\ 
  Scream of Stone (Schrei aus Stein) (1991) & Drama & 0.2445 \\ 
  Frank and Ollie (1995) & Documentary & 0.2445 \\ 
  Bye-Bye (1995) & Drama & 0.2445 \\ 
  Tigrero: A Film That Was Never Made (1994) & Documentary$|$Drama & 0.2445 \\ 
  Wend Kuuni (God's Gift) (1982) & Drama & 0.2445 \\ 
  Sonic Outlaws (1995) & Documentary & 0.2445 \\ 
  Getting Away With Murder (1996) & Comedy & 0.2445 \\ 
  Fausto (1993) & Comedy & 0.2445 \\ 
  Brothers in Trouble (1995) & Drama & 0.2445 \\ 
  Foreign Student (1994) & Drama & 0.2445 \\ 
  Tough and Deadly (1995) & Action$|$Drama$|$Thriller & 0.2445 \\ 
  Moonlight Murder (1936) & Mystery & 0.2445 \\ 
  Schlafes Bruder (Brother of Sleep) (1995) & Drama & 0.2139 \\ 
  Metisse (Cafe au Lait) (1993) & Comedy & 0.2139 \\ 
  Promise, The (Versprechen, Das) (1994) & Romance & 0.2139 \\ 
  Und keiner weint mir nach (1996) & Drama$|$Romance & 0.2139 \\ 
  Hungarian Fairy Tale, A (1987) & Fantasy & 0.2139 \\ 
  Liebelei (1933) & Romance & 0.2139 \\ 
  Paris, France (1993) & Comedy & 0.2139 \\ 
  Girl in the Cadillac (1995) & Drama & 0.2139 \\ 
  Hostile Intentions (1994) & Action$|$Drama$|$Thriller & 0.2139 \\ 
  Two Bits (1995) & Drama & 0.2139 \\ 
  Rent-a-Kid (1995) & Comedy & 0.2139 \\ 
  Beyond Bedlam (1993) & Drama$|$Horror & 0.2139 \\ 
  Touki Bouki (Journey of the Hyena) (1973) & Drama & 0.2139 \\ 
  Convent, The (Convento, O) (1995) & Drama & 0.2139 \\ 
  Open Season (1996) & Comedy & 0.2139 \\ 
  Lotto Land (1995) & Drama & 0.1834 \\ 
  Frisk (1995) & Drama & 0.1834 \\ 
  Shadow of Angels (Schatten der Engel) (1976) & Drama & 0.1834 \\ 
  Yankee Zulu (1994) & Comedy$|$Drama & 0.1834 \\ 
  Last of the High Kings, The (1996) & Drama & 0.1834 \\ 
  Sunset Park (1996) & Drama & 0.1834 \\ 
  Happy Weekend (1996) & Comedy & 0.1834 \\ 
  Criminals (1996) & Documentary & 0.1834 \\ 
  Happiness Is in the Field (1995) & Comedy & 0.1528 \\ 
  Associate, The (L'Associe)(1982) & Comedy & 0.1528 \\ 
  Target (1995) & Action$|$Drama & 0.1528 \\ 
  Relative Fear (1994) & Horror$|$Thriller & 0.1528 \\ 
  Honigmond (1996) & Comedy & 0.1528 \\ 
  Eye of Vichy, The (Oeil de Vichy, L') (1993) & Documentary & 0.1528 \\ 
  Sweet Nothing (1995) & Drama & 0.1528 \\ 
  Harlem (1993) & Drama & 0.1528 \\ 
  Condition Red (1995) & Action$|$Drama$|$Thriller & 0.1528 \\ 
  Homage (1995) & Drama & 0.1528 \\ 
  Superweib, Das (1996) & Comedy & 0.1222 \\ 
  Halfmoon (Paul Bowles - Halbmond) (1995) & Drama & 0.1222 \\ 
  Silence of the Palace, The (Saimt el Qusur) (1994) & Drama & 0.1222 \\ 
  Headless Body in Topless Bar (1995) & Comedy & 0.1222 \\ 
  Rude (1995) & Drama & 0.1222 \\ 
  Garcu, Le (1995) & Drama & 0.1222 \\ 
  Guardian Angel (1994) & Action$|$Drama$|$Thriller & 0.1222 \\ 
  Roula (1995) & Drama & 0.0917 \\ 
  Jar, The (Khomreh) (1992) & Drama & 0.0917 \\ 
  Small Faces (1995) & Drama & 0.0917 \\ 
  New York Cop (1996) & Action$|$Crime & 0.0917 \\ 
  Century (1993) & Drama & 0.0917 \\ 
\bottomrule\\
\end{supertabular}
\end{center}

}
\clearpage
\putbib[BIB1]
\end{bibunit}

\end{document}